\pdfoutput=1
\PassOptionsToPackage{numbers,compress}{natbib}
\documentclass{article}

\usepackage[main,final]{neurips_2026}

\usepackage[utf8]{inputenc}
\usepackage[T1]{fontenc}
\usepackage{amsmath,amssymb,amsfonts,amsthm,bm}

\usepackage{booktabs,multirow,tabularx,array}
\usepackage{graphicx}
\usepackage{subcaption}
\usepackage{wrapfig}
\usepackage{float}
\usepackage{placeins}
\usepackage[table]{xcolor}
\usepackage{enumitem}
\usepackage{needspace}
\usepackage[hypertexnames=false]{hyperref}
\usepackage{url}
\usepackage[nameinlink,noabbrev]{cleveref}
\usepackage{algorithm}
\usepackage{algpseudocode}
\usepackage{siunitx}
\usepackage{nicefrac}
\usepackage{microtype}

\hypersetup{
  pdftitle={HaM-World: Soft-Hamiltonian World Models with Selective Memory for Planning},
  pdfauthor={Haoyun Tang, Haodong Cui, Keyao Xu, Zhandong Mei, Kun Wang},
  colorlinks=true,
  linkcolor=blue!60!black,
  citecolor=blue!60!black,
  urlcolor=blue!60!black
}

\definecolor{TableHead}{HTML}{F3F6FA}
\newcommand{\headrow}{\rowcolor{TableHead}}

\newcommand{\method}{HaM-World}
\newcommand{\methodabbr}{HMW}
\newcommand{\baselineppo}{PPO}
\newcommand{\baselinesac}{SAC}
\newcommand{\baselinetdmpc}{TD-MPC2}
\newcommand{\baselinedreamer}{DreamerV3}

\newcommand{\z}{\mathbf{z}}
\newcommand{\qv}{\mathbf{q}}
\newcommand{\pv}{\mathbf{p}}
\newcommand{\cv}{\mathbf{c}}
\newcommand{\av}{\mathbf{a}}
\newcommand{\hv}{\mathbf{h}}
\newcommand{\sv}{\mathbf{s}}
\newcommand{\Loss}{\mathcal{L}}
\newcommand{\Ham}{\mathcal{H}}

\title{\textbf{\method}: Soft-Hamiltonian World Models with Selective Memory for Planning}

\author{%
  \textbf{Haoyun Tang}\textnormal{\textsuperscript{1}}\thanks{Equal contribution.}
  \quad \textbf{Haodong Cui}\textnormal{\textsuperscript{2}}\footnotemark[1]
  \quad \textbf{Keyao Xu}\textnormal{\textsuperscript{3}} \\
  \textbf{Zhandong Mei}\textnormal{\textsuperscript{1}}\thanks{Corresponding authors: Zhandong Mei (\texttt{zhdmei@mail.xjtu.edu.cn}) and Kun Wang (\texttt{wang.kun@ntu.edu.sg}).}
  \quad \textbf{Kun Wang}\textnormal{\textsuperscript{4}}\footnotemark[2] \\
  \textnormal{\textsuperscript{1}Xi'an Jiaotong University}
  \quad \textnormal{\textsuperscript{2}Huazhong University of Science and Technology} \\
  \textnormal{\textsuperscript{3}Nankai University}
  \quad \textnormal{\textsuperscript{4}Nanyang Technological University}
}

\date{}

\begin{document}

\maketitle

\begin{abstract}
World models enable model-based planning through learned latent dynamics, but imagined rollouts become unstable as the planning horizon grows or the dynamics distribution shifts: an unconstrained predictor contracts, because each predicted state is fed back as the next input, so the latent distribution shrinks over the rollout until distinct states merge. We argue that this drift and contraction reflect two missing structures in planner-facing latents: a history-conditioned predictive state that summarizes past observations and actions, and geometric organization that separates configuration, momentum, and task semantics. We propose \method{} (\methodabbr{}), a structured world model that decomposes the latent state into a canonical-like $(\qv,\pv)$ subspace and a context subspace $\cv$, while using Mamba selective state-space memory as the history-conditioned input to the same latent dynamics. Within this interface, $(\qv,\pv)$ evolves through an energy-derived Hamiltonian vector field plus learnable residual/control dynamics, while $\cv$ captures semantic, dissipative, and non-conservative factors. This gives the planner a single latent state shared by dynamics prediction, reward/value estimation, imagined rollouts, and CEM action search. On six DeepMind Control Suite tasks, \method{} ranks first on four and second on the remaining two, with the highest Avg.\ AUC over all six tasks ($91.9$, $+10.3\%$ over \baselinetdmpc{}). Within the horizon range the planner actually unrolls, it reduces imagined-rollout error to $45\%$ of \baselinetdmpc{}'s and wins 11/12 $k\!\in\!\{3,5,7\}$ MSE cells. Under 12 OOD perturbations spanning dynamics shifts, action delay, and observation masking it attains the highest return in every condition ($+10.2\%$ on Finger Spin, $+13.6\%$ on Reacher Easy). Ablation studies show that both selective memory and the soft Hamiltonian prior contribute to improved model performance. Mechanism diagnostics support the intended design, showing bounded action-free energy drift, structured energy variation under policy rollouts, and coherent control-induced energy transfer.
Code: \url{https://github.com/HaoyunT/HaM\_World}.
\end{abstract}
\section{Introduction}

In recent years, world models have demonstrated strong capability in representing real-world dynamics through structured latent spaces~\citep{ha2018recurrent,hafner2025mastering}. However, as the planning horizon increases or the dynamics shift, errors in imagined rollouts accumulate rapidly, leading to instability in long-horizon planning and poor physical-structure extrapolation under perturbed control regimes. This limitation arises because existing approaches primarily capture statistical correlations rather than the underlying generative structure of the environment. Rather than relying solely on correlation fitting, an effective world model should incorporate structure-aware inductive biases that reflect physical regularities, such as invariances and conservation laws~\citep{greydanus2019hamiltonian,Zhong2020Symplectic}. In particular, a monolithic latent representation entangles configuration, momentum, and task semantics, causing multi-step prediction errors to grow without structural constraints during planning. An unconstrained predictor also contracts: each imagined state is fed back as the next input, so a small per-step contraction compounds over the planning horizon and the latent distribution narrows until distinct states merge. A Hamiltonian vector field is divergence-free and therefore neither contracts nor expands the distribution to first order, which is why geometric structure is a candidate remedy rather than more capacity~\citep{janner2019trust,pmlr-v162-wang22c,liu2023learning}.

We view this limitation as stemming from two missing pieces in planner-facing latent dynamics. \textbf{On the input side}, partial observability and action delay make a single-frame latent insufficient to determine the next transition; selective sequence models address this through input-dependent memory~\citep{gu2024mamba}. \textbf{On the output side}, a unified latent variable entangles configuration, velocity trends, and task semantics, leaving multi-step rollouts without structure-consistent constraints. Physics priors and predictive representations attack related issues from complementary angles~\citep{troch2025action,assran2023self}. These two sides affect different stages of the same latent dynamics, and strengthening either one alone is unlikely to make the planner improve simultaneously in training return, long-horizon rollout consistency, and physical-structure extrapolation under distribution shifts and perturbations.

Our central view is that \textit{a planning-oriented world model should treat history-conditioned state sufficiency and geometric consistency as two coupled stages of unified latent dynamics}. Memory supplies the conditioning that lets the latent dynamics predict the next transition from a summary of the past rather than a single frame, upon which Soft-Hamiltonian structure in $(\qv,\pv)$ constrains error accumulation over the horizons the planner actually unrolls. The structure does three things for the planner that correlation fitting does not: it counteracts contraction of the latent distribution under repeated prediction; it organizes rollouts around a learned scalar energy $H_\phi$, so that energy changes are accounted for by action work rather than by unconstrained drift; and it makes the dynamics partly inspectable, because the learned energy field, its contours, and their coupling to the control signal can be plotted and tested rather than only measured by return. Based on this perspective, we propose \method{} (\methodabbr{}), whose unified latent interface $\z_t{=}[\qv_t,\pv_t,\cv_t]$ is shared across dynamics prediction, reward/value estimation, imagined rollouts, and cross-entropy method (CEM) planning~\citep{rubinstein1999cross}. In the canonical $(\qv,\pv)$ subspace, the Hamiltonian flow field supplies energy-organized local directions, while learnable residual/control dynamics capture controlled deviations; $\cv$ captures semantic and non-conservative factors relevant to reward prediction, value estimation, and action search. Mamba selective memory~\citep{gu2024mamba} provides history-conditioned inputs to the same latent dynamics, keeping the planner interface unified across training, imagined rollout, and CEM search without a separate recurrent rollout state or planner-specific latent state.

\paragraph{Contributions.}
(i) \textbf{Structured planner-facing dynamics.} We propose \method{}, a q/p/c latent factorization that couples Mamba selective memory~\citep{gu2024mamba} with Soft-Hamiltonian $(\qv,\pv)$ dynamics, giving CEM a single latent interface shared by prediction, reward/value estimation, and action search, and exposing a single learned energy $H_\phi$ that makes the planner-facing dynamics inspectable;
(ii) \textbf{Mechanism-aware evaluation.} We evaluate return, rollout MSE for $k$ up to $20$, 12 out-of-distribution (OOD) perturbations, and Hamiltonian mechanism analysis on six DeepMind Control Suite (DMC) tasks~\citep{tassa2018deepmind};
(iii) \textbf{Empirical performance.} \method{} reaches the highest Avg.\ AUC over six tasks ($91.9$, $+10.3\%$), reduces rollout error \emph{within the planning-relevant range} to $45\%$ of a strong baseline, wins 11/12 $k\!\in\!\{3,5,7\}$ MSE cells, and obtains the highest absolute return in every OOD condition, with $10.2\%$ and $13.6\%$ average OOD-return gains on Finger Spin and Reacher Easy. Ablations show that selective memory and the Soft-Hamiltonian prior both contribute to improved performance.

\section{Related Work}\label{sec:related}

\paragraph{World models, predictive latents, and factorized state.}
Model-based reinforcement learning learns dynamics that can be queried by a planner or behavior optimizer, from Dyna~\citep{sutton1991dyna} and World Models~\citep{ha2018recurrent} to PlaNet/Dreamer~\citep{hafner2019learning,Hafner2020Dream,hafner2021mastering,hafner2025mastering}, MuZero~\citep{schrittwieser2020mastering}, TD-MPC~\citep{pmlr-v162-hansen22a,hansen2024tdmpc}, MBPO~\citep{janner2019trust}, and recent planner-policy, multi-task, and scalable MBRL variants~\citep{gumbsch2023learning,ICLR2025_8636419d,georgiev2025pwm,wang2024parallelizing,sukhija2026sombrl}. Model-free actor-critic methods such as PPO and SAC remain strong control references~\citep{schulman2017proximal,haarnoja2018soft}, but they do not expose latent rollouts. Transformers~\citep{micheli2023transformers,zhang2023storm,wu2024ivideogpt}, diffusion world models~\citep{alonso2024diffusion,lee2026edeline}, large-scale generative or multimodal environments~\citep{bruce2024genie,yang2024learning,mazzaglia2024genrl}, operator-learning views~\citep{novelli2024operator}, and policy-shaped prediction~\citep{hutson2024policy} expand the transition-model design space; the taxonomy in \citet{ding2025understanding} places these families in a broader world-model landscape. Reconstruction-free learning further replaces pixel reconstruction with objectives closer to prediction and decision making: JEPA-style models~\citep{assran2023self,bardes2024revisiting,mo2024connecting,assran2025v}, invariant or control-oriented representations~\citep{zhang2021learning}, Dreaming/DreamerPro~\citep{okada2021dreaming,deng2022dreamerpro}, and TD-JEPA~\citep{bagatella2026tdjepa} all suggest that representation objectives should be aligned with downstream behavior. Orthogonally, latent factorization has been used to improve control and generalization, including SOLAR~\citep{zhang2019solar}, Denoised MDPs~\citep{pmlr-v162-wang22c}, IFactor~\citep{liu2023learning}, Plan2Explore~\citep{sekar2020planning}, and recent semantic/dynamic or long-horizon decompositions~\citep{wang2025disentangled,zhang2026dymodreamer,roder2026dynamicsaligned,wang2026dmwm}. \method{} combines these threads in a narrower planner-facing form: the latent is not only predictive, but explicitly split into coordinates used by reward/value estimation, imagined rollouts, and CEM action search throughout training and evaluation.

\paragraph{Physics priors and memory for long-horizon rollouts.}
Hamiltonian Neural Networks~\citep{greydanus2019hamiltonian}, Symplectic ODE-Net~\citep{Zhong2020Symplectic}, action-conditioned Hamiltonian models~\citep{troch2025action}, stable port-Hamiltonian networks~\citep{roth2026stable}, and embodied physics-prior world models~\citep{shang2026roboscape} show that energy-based geometric structure can improve stability, plausibility, and extrapolation. Strict conservation is nevertheless too restrictive for planning in contact-rich controlled systems, where friction, actuation, task semantics, observation noise, and reward-relevant context create non-conservative effects. \method{} therefore uses a Soft-Hamiltonian prior on $(\qv,\pv)$: the Hamiltonian vector field gives a structured backbone, residual/control dynamics absorb controlled deviations, and $\cv$ carries semantic and dissipative context. This differs from simply adding smoothness constraints such as spectral normalization~\citep{miyato2018spectral}, because the prior is placed on the state actually queried by the planner. On the input side, partial observability makes a single observation insufficient to determine the next transition, so history-dependent representations are necessary; recurrent state-space models, Transformers, and state-space sequence models (SSMs) address this through learned memory~\citep{hafner2019learning,micheli2023transformers,lu2023structured,lv2024decision,aljalbout2025accelerating}. For planner rollouts, however, the memory must be both efficient over long horizons and adaptive to the current action-conditioned transition: RNN/GRU memory offers a compact recurrent state with input-dependent gates, while Transformer memory can model rich context at higher sequence cost. Structured SSMs provide recurrent long-context filtering for RL, and Mamba further makes the state-space update input-selective~\citep{lu2023structured,gu2024mamba,pmlr-v235-dao24a,lv2024decision,aljalbout2025accelerating}. We therefore use Mamba as a history-conditioned input to the same latent dynamics, forming a history-conditioned predictive state within the planner interface.

\section{Preliminaries and Problem Setup}\label{sec:prelim}

We consider continuous-control trajectories $\tau=\{(o_t,\av_t,r_t)\}_{t=0}^{T}$ with observations $o_t$, actions $\av_t$, and rewards $r_t$. A planner-facing world model maps observations to latents and rolls them forward under candidate actions over a finite planning horizon,
\begin{equation}
\z_t=E_\phi(o_t),\qquad
(\hat\z_{t+1},\hv_{t+1})=F_\phi(\hat\z_t,\av_t,\hv_t),\qquad
\hat r_t=R_\phi(\hat\z_t,\av_t),\quad \hat V_t=V_\phi(\hat\z_t).
\end{equation}
Here $E_\phi,F_\phi,R_\phi,V_\phi$ are the encoder, transition, reward, and value heads; $\hat{\z}_t$ denotes the model-predicted latent for imagined rollouts, and $\hv_t$ is the history state. At decision time, CEM searches action sequences by maximizing model-imagined return from predicted rewards and terminal value,
\begin{equation}\label{eq:cem-objective}
\av_{t:t+H-1}^{\star}
\in \arg\max_{\av_{t:t+H-1}}
\sum_{k=0}^{H-1}\gamma^k \hat r_{t+k}
\;+\;\gamma^H \hat V_\phi(\hat\z_{t+H}),
\quad
\hat\z_{t+k+1}=F_\phi(\hat\z_{t+k},\av_{t+k},\hv_{t+k}).
\end{equation}
The central design question is therefore not only how to learn an accurate one-step predictor, but how to choose a latent state $\z_t$ whose repeated rollout remains stable under partial observations and directly useful for reward/value prediction and action search.

\paragraph{The planner-facing state.}
Strict predictive sufficiency would require a statistic $s_t=\psi(\mathcal H_t)$ of the history $\mathcal H_t=(o_{\le t},\av_{<t})$ with $p(o_{t+1},r_t\mid \mathcal H_t,\av_t)=p(o_{t+1},r_t\mid s_t,\av_t)$. We use $s_t=(\z_t,\hv_{t-1})$ as the planner-facing state, where $\z_t=E_\phi(o_t)$ encodes the current observation and $\hv_t=\mathrm{Memory}_\phi(\z_t,\av_t,\hv_{t-1})$ is a fixed-size summary of the past that conditions the transition. The no-memory and GRU ablations (Section~\ref{sec:ablation}) and the multi-step rollout and reward errors test whether this state retains what finite-horizon planning needs. ``History-conditioned'' refers throughout to this construction.

\paragraph{What $(\qv,\pv)$ and $H$ mean in a learned latent space.}
Classical mechanics describes a system by its configuration $q$ and conjugate momentum $p$. One scalar $H(q,p)$, the energy, determines both updates: $\dot q=\partial_pH$ says position changes at a rate set by momentum, and $\dot p=-\partial_qH$ says momentum changes with the energy slope along position, so motion follows energy contours instead of drifting --- the property we want when a rollout is fed back into itself many times. In a learned latent space, $\qv$ and $\pv$ are \emph{learned} canonical-like coordinate blocks rather than dimension-wise physical quantities, $H_\phi$ is an energy-like scalar field rather than identified mechanical energy, and we borrow only the \emph{directions} of the flow: $\xi_H=(\partial_{\pv}H_\phi,-\partial_{\qv}H_\phi)$ satisfies $\nabla H_\phi^\top\xi_H=0$, so moving along it leaves the learned energy unchanged to first order --- exactly what repeated unrolling needs. Strict conservation is neither required nor desirable here: the learned residual fits dissipative dynamics, the control drive injects action work, and $\cv$ carries non-Hamiltonian information. Appendix~\ref{app:qp-meaning} expands on this reading.

\section{Method}

\begin{figure}[H]
  \centering
  \includegraphics[width=\linewidth,height=0.42\textheight,keepaspectratio]{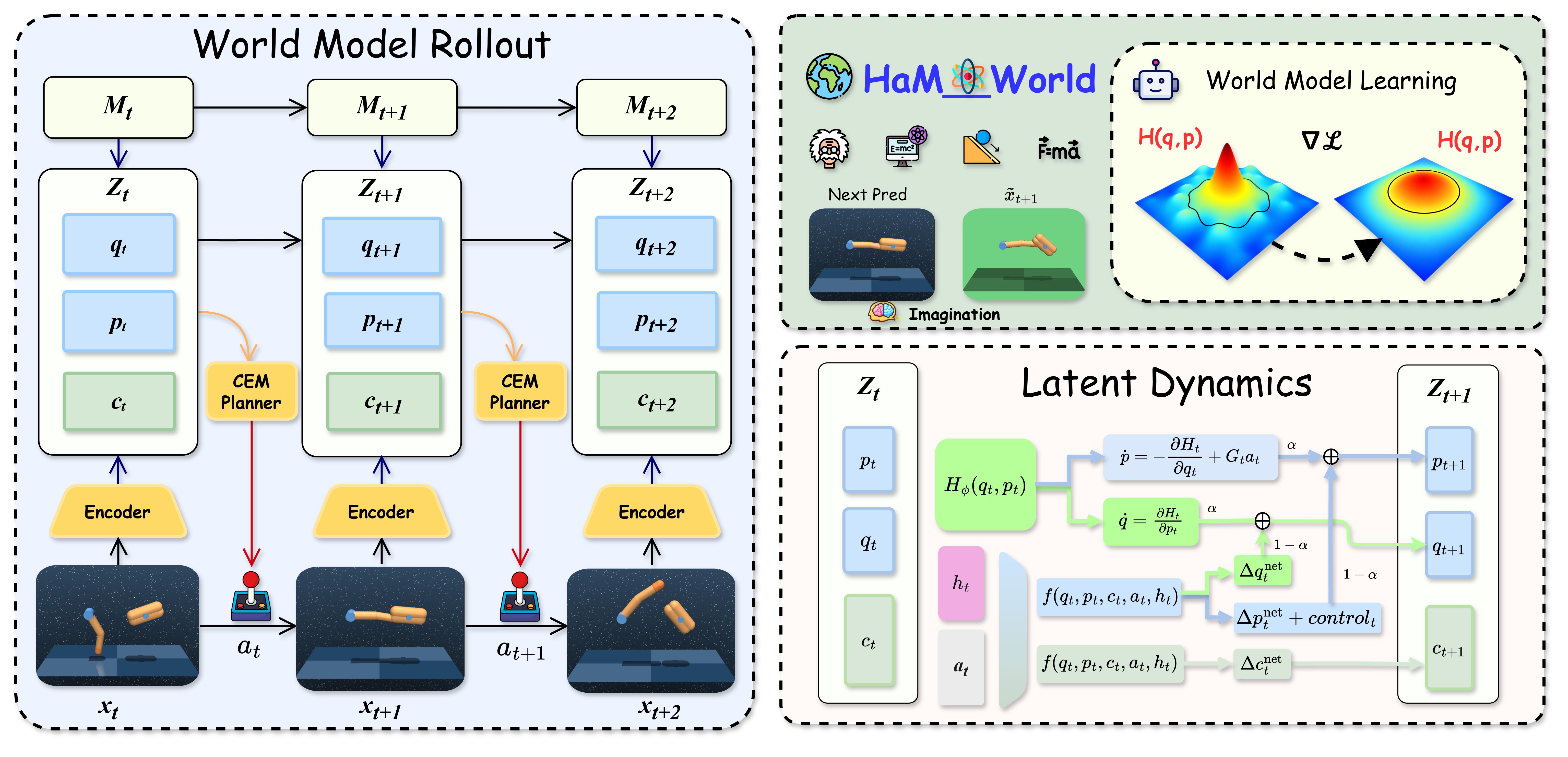}
  \caption{Architecture of \method{}. Observations are encoded into Hamiltonian state $(\qv_t,\pv_t)$, context $\cv_t$, and Mamba memory $\hv_t$; a shared latent transition supports reward/value prediction and CEM planning inside one unified planner-facing interface.}
  \label{fig:method}
\end{figure}

\method{} instantiates the planner-facing world model above with a q/p/c latent split and a single shared dynamics interface. Actions are searched by CEM inside the learned dynamics. The overall architecture is shown in Figure~\ref{fig:method}.

\subsection{q/p/c-Structured Latents and Soft-Hamiltonian Pair Dynamics}\label{sec:method-structured}

Long-horizon planning asks the same latent dynamics to satisfy two requirements. On one hand, repeated planner rollouts need a stable geometric backbone. On the other, real control tasks contain contact, friction, reward semantics, and partial observability. If all of these factors are compressed into a single unstructured latent, the model can oscillate between short-term fitting and long-term rollout stability. \method{} therefore writes the encoder output as
$\z_t = E_\phi(o_t) = [\qv_t,\pv_t,\cv_t]$: $(\qv,\pv)$ form a canonical subspace and $\cv$ forms a context subspace. The reward head, value head, imagined rollout, and CEM planner all share the same $\z_t$, so this division of roles acts directly on the state queried by the planner, rather than as a training-only or post-hoc explanatory variable. The Soft-Hamiltonian pair dynamics balance structural stability and expressive flexibility. A generic MLP latent dynamics lacks energy-field constraints, whereas a strictly Hamiltonian model over-constrains dissipative controlled dynamics. We therefore allow only $(\qv,\pv)$ to form the main dynamics backbone through a learnable energy field $\Ham=H_\phi(\qv,\pv)$, while augmenting that pair with residual/control dynamics and using $\cv$ to represent complementary semantic and non-conservative context. The controlled Hamiltonian equations
$\dot\qv = \partial\Ham/\partial\pv,\ \dot\pv = -\partial\Ham/\partial\qv + g(\qv)\av$~\citep{Zhong2020Symplectic,troch2025action}
decompose internal energy geometry from action work; \method{} discretizes this structure and embeds it into the planner-facing dynamics used by CEM during every imagined rollout within the shared transition interface.

\paragraph{Why an energy-organized field.}
The difficulty is repeated self-iteration: to score one candidate the planner feeds the model's own output back in for $H$ steps, and an unconstrained transition can then \emph{collapse} trajectories into a low-dimensional subspace, making candidates indistinguishable to the reward/value heads, or \emph{expand} them outside the training region, where those heads score unreliably. The raw Hamiltonian field addresses exactly these two modes: it is orthogonal to the energy gradient and divergence-free, so in continuous time it neither changes $H$ nor compresses or expands phase-space volume, and a single $H_\phi$ drives both updates so the coordinates are jointly constrained (Appendix~\ref{app:proof-energy}). With a residual, control drive, context channel, and forward Euler these survive only as a bias.

\paragraph{Hamiltonian-pair update.}
Given the Mamba memory output $\hv_t = \mathrm{Memory}_\phi(\z_t,\av_t,\hv_{t-1})$, the Soft-Hamiltonian update for the Hamiltonian pair produces
\begin{equation}\label{eq:soft-ham}
  \begin{aligned}
    \bigl[\Delta\qv^{\text{net}}_t,\Delta\pv^{\text{net}}_t\bigr] &= f_{\text{core}}(\qv_t,\pv_t,\cv_t,\av_t,\hv_t),\quad
    \Ham_t = H_\phi(\qv_t,\pv_t),\quad
    \mathbf{G}_t = G_\phi(\qv_t,\pv_t,\cv_t,\av_t,\hv_t), \\
    \Delta\qv_t &= (1-\alpha)\,\Delta\qv^{\text{net}}_t + \alpha\,\partial_{\pv}\Ham_t,\quad
    \Delta\pv_t = (1-\alpha)\,\Delta\pv^{\text{net}}_t - \alpha\,\partial_{\qv}\Ham_t + \mathbf{G}_t\av_t,
  \end{aligned}
\end{equation}
followed by $\qv_{t+1}{=}\qv_t{+}\Delta\qv_t$ and $\pv_{t+1}{=}\pv_t{+}\Delta\pv_t$. The term ``Soft-Hamiltonian'' reflects two design choices: $\alpha$ mixes the action-conditioned network update with the Hamiltonian vector field on the canonical pair, while $\mathbf{G}_t\av_t$ adds an explicit learned control drive to the momentum update. As $\alpha{\to}0$, the pair update degenerates to a data-driven update plus control; as $\alpha{\to}1$, the canonical part approaches a controlled Hamiltonian update with the same control channel. Intermediate $\alpha$ keeps action-free energy drift small when the network update is aligned with the Hamiltonian direction, while retaining the ability to fit contact, friction, and task objectives during closed-loop planning across horizons and perturbations seen by the planner during model-predictive control rather than only one-step prediction in isolation during training.

Eq.~\eqref{eq:soft-ham} describes the update before Hamiltonian-gradient clipping and latent norm projection; Appendix~\ref{app:integrator} specifies these safeguards.

\paragraph{Context and selective memory.}
The context variable $\cv$ is introduced because not all control-relevant information should obey the Hamiltonian energy field. Forcing contact switches, friction, task semantics, or observation noise into $(\qv,\pv)$ can damage the stabilizing role of the energy field; discarding them would harm reward/value prediction and planning. The context update is
$\Delta\cv_t = f_c(\qv_t,\pv_t,\cv_t,\av_t,\hv_t)$, so $\cv$ remains available within the unified latent dynamics as complementary information for these non-Hamiltonian factors. ``Context'' names an architectural role: a channel not required to follow the vector field of $H_\phi$. The Mamba selective scan~\citep{gu2024mamba,pmlr-v235-dao24a} outputs $\hv_t$ only as a \textit{history-conditioned input} to the same latent dynamics, entering $f_{\text{core}}$, $G_\phi$, and $f_c$ simultaneously. The model has one planner-facing latent dynamics; memory supplies the missing history information under partial observability, action delay, and long horizons during imagined search.

\subsection{Mechanism-Oriented Stability Analysis}\label{sec:theory}

We package the mechanism as a local statement about the map the planner actually iterates, together with the diagnostics it predicts. Accumulated prediction error over multiple steps is characterized empirically by the multi-horizon rollout errors in Section~\ref{sec:exp_main} and Appendix~\ref{app:long-horizon}.

\vspace{2pt}
\begin{center}
\setlength{\fboxsep}{4pt}
\noindent\fbox{%
\begin{minipage}{0.965\linewidth}
\small
\textbf{Soft-Hamiltonian Mechanism.}
The planner unrolls the same transition used in training, including memory and clipping safeguards (Appendix~\ref{app:integrator}). On steps without Hamiltonian-gradient clipping or latent norm projection, this is the unroll of Eq.~\eqref{eq:soft-ham} (Appendix~\ref{app:proof-unroll}). The raw Hamiltonian field is energy-orthogonal and divergence-free on the canonical pair (Eq.~\eqref{eq:symplectic-cancel}); on such steps, control and the alignment residual are the only first-order sources of energy change. Active clipping or projection can add a first-order contribution (Appendix~\ref{app:proof-energy}). The corresponding diagnostics are slow action-free energy drift and push-dependent contour crossing, tested empirically in Section~\ref{ssec:mech} and Appendix~\ref{app:diagnostics}.
\end{minipage}}
\end{center}
\vspace{-2pt}

\subsection{Training Objective and Planning}

The training objectives are designed to approximate the desirable behaviors characterized in Section~\ref{sec:theory}, rather than enforcing them as strict constraints. The total loss is
\begin{equation}\label{eq:total-loss}
\Loss_{\text{total}}=\Loss_{\text{repr}}+\beta_{\text{dyn}}\Loss_{\text{dyn}}+\beta_{\text{roll}}\Loss_{\text{roll}}+\beta_{\text{r}}\Loss_{\text{reward}}+\beta_{\text{v}}\Loss_{\text{value}}+\beta_{\text{p}}\Loss_{\text{policy}}+\beta_{\text{geo}}\Loss_{\text{geo}},
\end{equation}
where $\Loss_{\text{repr}}$ is a JEPA-style online/EMA latent alignment objective~\citep{assran2023self,bardes2024revisiting}, $\Loss_{\text{dyn}}$ is a one-step latent consistency loss, reward/value use DreamerV3-style symlog two-hot regression with EMA value targets~\citep{hafner2025mastering}, and $\Loss_{\text{policy}}$ trains the CEM warm-start prior by behavior cloning. Below we highlight three representative losses that are directly aligned with the Soft-Hamiltonian structure; all other terms and weights are in Appendix~\ref{app:implementation}.

\textbf{Multi-step rollout consistency.}
\begin{equation}\label{eq:roll-main}
\Loss_{\text{roll}} = \frac{1}{|\mathcal{S}|}\sum_{s\in\mathcal{S}}\sum_{k=1}^{K}\bigl\|\hat{\z}_{s+k}^{(s)} - \mathrm{sg}(\z_{s+k})\bigr\|_2^2,
\end{equation}
where $\hat{\z}_{s+k}^{(s)}$ is the $k$-step prediction rolled out from $\z_s$ with $\hv_s$ as the initial memory condition. This term provides a discrete surrogate for controlling multi-step error accumulation, directly constraining the rollout horizon relevant for planning.

\textbf{Hamiltonian alignment.}
\begin{equation}\label{eq:ham-align}
\Loss_{\text{ham}} = \bigl\|\Delta\qv_t^{\mathrm{net}} - \partial_{\pv}\Ham_t\bigr\|_2^2 + \bigl\|\Delta\pv_t^{\mathrm{net}} + \partial_{\qv}\Ham_t\bigr\|_2^2.
\end{equation}
This loss acts as a soft directional bias, aligning the network branch with the Hamiltonian vector field while leaving the learned control drive outside the alignment residual.

\textbf{Action-free energy regularization.}
\begin{equation}\label{eq:energy-main}
\Loss_{\text{energy}} = \mathbb{E}_{t:\|\av_t\|<\epsilon}\bigl[(\Ham(\qv_{t+1},\pv_{t+1}) - \Ham(\qv_t,\pv_t))^2\bigr].
\end{equation}
This term directly suppresses energy drift in the action-free setting, encouraging the stable behavior described in Section~\ref{sec:theory} for passive rollouts.

The geometric regularization term is defined as
\[
\Loss_{\text{geo}} = \lambda_{\mathrm{ham}}\Loss_{\text{ham}} + \lambda_{\mathrm{en}}\Loss_{\text{energy}} + \lambda_{\mathrm{sa}}\Loss_{\text{sa}} + \lambda_{\mathrm{temp}}\Loss_{\text{temp}} + \lambda_{\mathrm{dec}}\Loss_{\text{dec}} + \lambda_{\mathrm{c}}\Loss_{\text{c}} .
\]
Full formulas and weights are provided in Appendix~\ref{app:implementation}. The planner performs horizon-$H$ CEM optimization over $(\z_t,\hv_t)$, searching action sequences using the shared latent dynamics, reward head, and value head.

\section{Experimental Setup}\label{sec:experimental_setup}

We evaluate on a four-task core suite --- Reacher Easy, Finger Spin, Cheetah Run, and Cartpole Swingup --- from the DeepMind Control Suite~\citep{tassa2018deepmind}, plus two additional tasks, Walker Run and Hopper Hop, chosen because the core suite underrepresents sustained coordination, momentum, and repeated contact. All six use state observations, 100k environment steps, and 3 seeds (7/8/9). We report final return, AUC, latent-rollout MSE for $k\!\in\!\{3,5,7\}$ in the main text and $k$ up to $20$ in Appendix~\ref{app:long-horizon}, and zero-shot OOD return. The OOD protocol uses 12 perturbations over Reacher Easy and Finger Spin, covering dynamics changes, action delay, and observation masking; full task and evaluation details are in Appendix~\ref{app:implementation}.

\paragraph{Baselines.}
We compare model-free actor-critic baselines (\baselineppo{}~\citep{schulman2017proximal}, \baselinesac{}~\citep{haarnoja2018soft}) with model-based world models (\baselinedreamer{}~\citep{hafner2025mastering}, \baselinetdmpc{}~\citep{hansen2024tdmpc}). PPO and SAC lack explicit latent rollouts, so they are omitted from latent-rollout MSE. Methods using external pretraining or visual encoders are excluded.

\section{Main Results}\label{sec:exp_main}

We present the evidence as a sequence of observations: standard-control return, planner-relevant rollout consistency, zero-shot OOD return, and finally the Hamiltonian mechanism diagnostics that connect the gains to the proposed memory-and-geometry design, summarized here and given in full in Appendix~\ref{app:diagnostics}.

\subsection{Control Performance and Long-Horizon Consistency}

\begin{figure}[!htbp]
  \centering
  \begin{subfigure}[t]{0.245\linewidth}\centering
    \includegraphics[width=\linewidth]{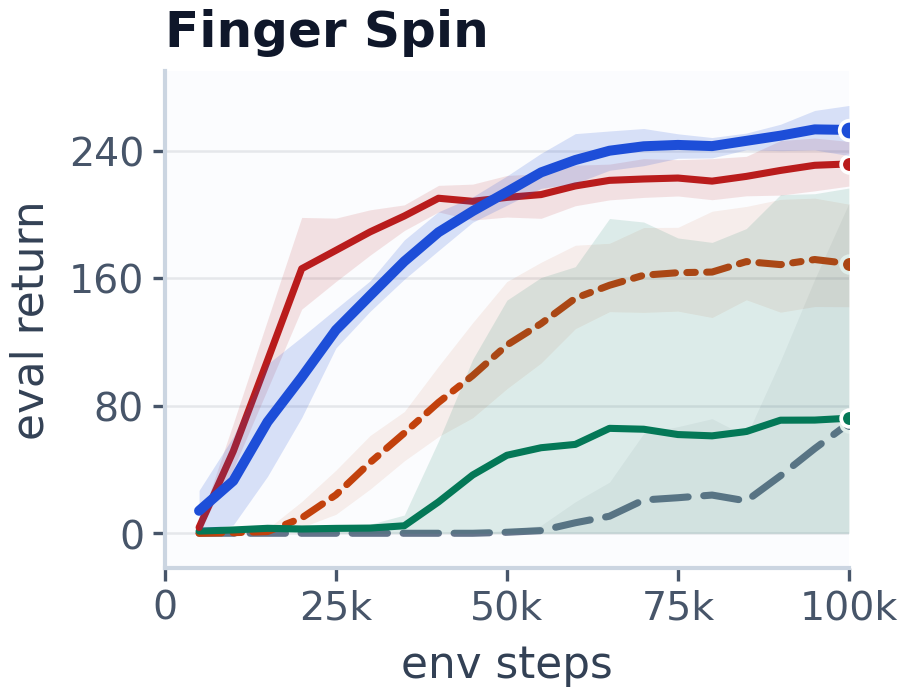}
    \label{fig:results-finger}
  \end{subfigure}\hfill
  \begin{subfigure}[t]{0.245\linewidth}\centering
    \includegraphics[width=\linewidth]{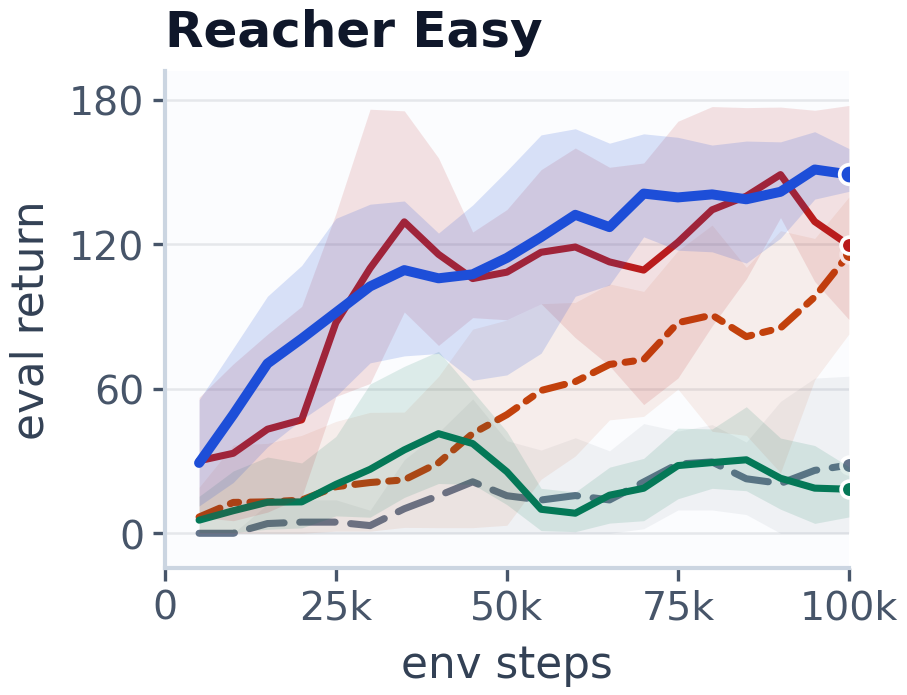}
    \label{fig:results-reacher}
  \end{subfigure}\hfill
  \begin{subfigure}[t]{0.245\linewidth}\centering
    \includegraphics[width=\linewidth]{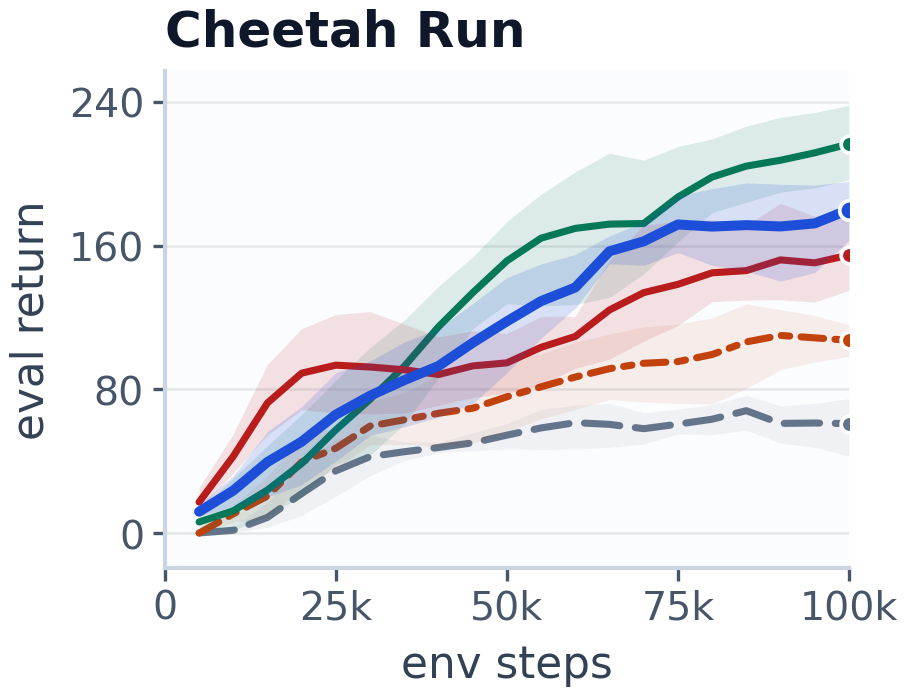}
    \label{fig:results-cheetah}
  \end{subfigure}\hfill
  \begin{subfigure}[t]{0.245\linewidth}\centering
    \includegraphics[width=\linewidth]{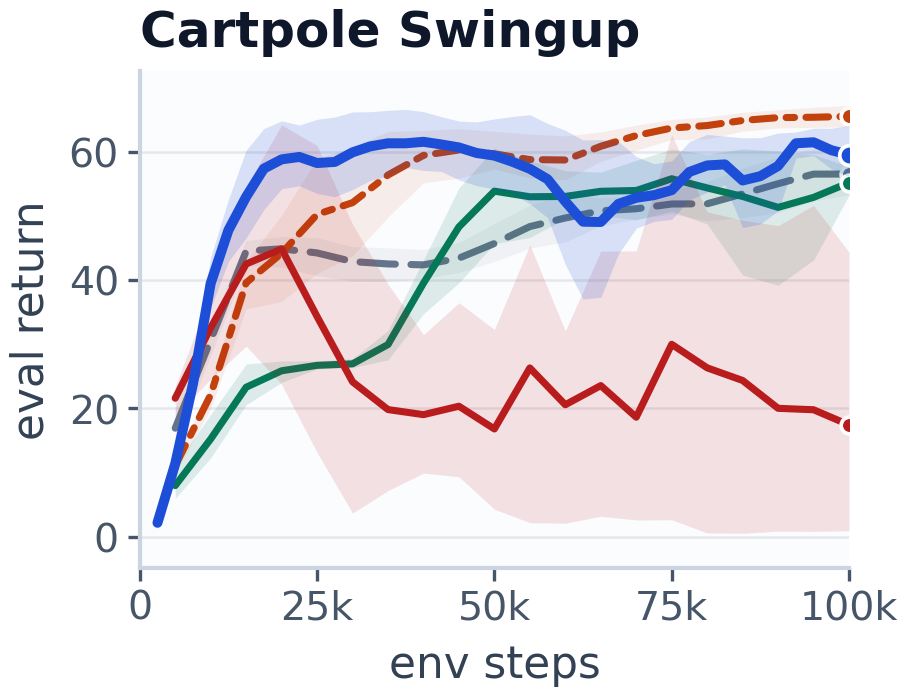}
    \label{fig:results-cartpole}
  \end{subfigure}\\[1pt]
  \includegraphics[width=0.5\linewidth]{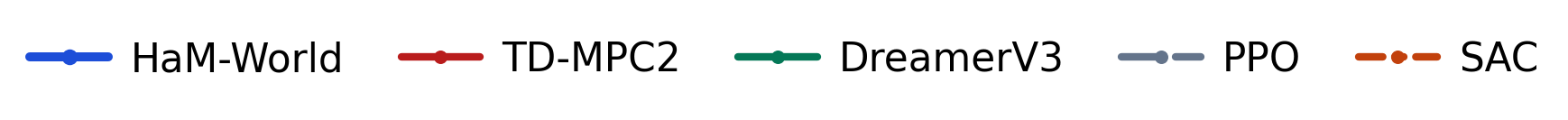}
  \caption{Return learning curves on the four core tasks (mean curve with seed envelope). \method{} obtains the highest final return on Finger and Reacher and the second-best on Cheetah and Cartpole; summarized by training AUC it obtains the highest average score. Final returns for Walker Run and Hopper Hop are in Appendix~\ref{app:extra-tasks}.}
  \label{fig:return-curves}
\end{figure}

Table~\ref{tab:overview} groups two core metrics by row: (I) final return at 100k environment steps and Avg.\ AUC; (II) $k\in\{3,5,7\}$-step imagined-rollout latent MSE. We report coordinate-summed latent error as the main latent-MSE metric because planning/reward/value heads consume the full latent vector, so total rollout deviation, rather than only per-coordinate drift, matters for imagined planning. \textbf{Bold}/\underline{underline} denote the best/second-best value in each column within each task block of the table.

\begin{table}[!htbp]
  \centering
  \caption{Main table (mean$\pm$std, 3 seeds): (I) final return and average AUC; (II) $k\in\{3,5,7\}$-step latent-rollout MSE, applicable only to model-based methods. Two further tasks (Walker Run, Hopper Hop) and the Avg.\ AUC over all six tasks are reported in Appendix~\ref{app:extra-tasks}, and $k$ up to $20$ in Appendix~\ref{app:long-horizon}.}
  \label{tab:overview}
  \scriptsize
  \setlength{\tabcolsep}{3pt}
  \renewcommand{\arraystretch}{1.1}
  \begin{tabular*}{\linewidth}{@{\extracolsep{\fill}} l ccc ccc ccc ccc c@{}}
    \toprule
    Method & \multicolumn{3}{c}{Finger Spin} & \multicolumn{3}{c}{Reacher Easy} & \multicolumn{3}{c}{Cheetah Run} & \multicolumn{3}{c}{Cartpole Swingup} & Avg.\ \\
    \cmidrule(lr){2-4}\cmidrule(lr){5-7}\cmidrule(lr){8-10}\cmidrule(lr){11-13}\cmidrule(lr){14-14}
    \rowcolor{TableHead}\multicolumn{14}{c}{\textit{(I) Final return @ 100k env steps $\uparrow$ \quad (Avg.\ = AUC over training, core suite)}} \\
    \midrule
    \baselineppo{}    & \multicolumn{3}{c}{$70.3{\scriptstyle\pm 99.5}$}   & \multicolumn{3}{c}{$18.9{\scriptstyle\pm 15.1}$}   & \multicolumn{3}{c}{$67.5{\scriptstyle\pm 17.4}$}        & \multicolumn{3}{c}{$56.8{\scriptstyle\pm 2.5}$} & $30.3$ \\
    \baselinesac{}    & \multicolumn{3}{c}{$173.6{\scriptstyle\pm 26.1}$}  & \multicolumn{3}{c}{$\underline{118.9}{\scriptstyle\pm 3.9}$} & \multicolumn{3}{c}{$108.5{\scriptstyle\pm 6.5}$}    & \multicolumn{3}{c}{$\mathbf{65.8}{\scriptstyle\pm 1.3}$} & $70.9$ \\
    \baselinedreamer{} & \multicolumn{3}{c}{$72.3{\scriptstyle\pm 102.3}$} & \multicolumn{3}{c}{$18.4{\scriptstyle\pm 7.9}$}    & \multicolumn{3}{c}{$\mathbf{216.4}{\scriptstyle\pm 17.0}$} & \multicolumn{3}{c}{$55.8{\scriptstyle\pm 0.4}$}      & $58.6$ \\
    \baselinetdmpc{}  & \multicolumn{3}{c}{$\underline{232.2}{\scriptstyle\pm 8.2}$} & \multicolumn{3}{c}{$105.1{\scriptstyle\pm 52.3}$} & \multicolumn{3}{c}{$155.7{\scriptstyle\pm 17.4}$}      & \multicolumn{3}{c}{$15.0{\scriptstyle\pm 17.0}$}     & $\underline{107.7}$ \\
    \method{}         & \multicolumn{3}{c}{$\mathbf{254.0}{\scriptstyle\pm 14.1}$} & \multicolumn{3}{c}{$\mathbf{150.6}{\scriptstyle\pm 7.5}$} & \multicolumn{3}{c}{$\underline{184.4}{\scriptstyle\pm 8.4}$} & \multicolumn{3}{c}{$\underline{58.9}{\scriptstyle\pm 3.8}$}      & $\mathbf{117.9}$ \\
    \midrule
    \rowcolor{TableHead}\multicolumn{14}{c}{\textit{(II) $k$-step latent rollout MSE $\downarrow$ \quad (model-based methods only)}} \\
    \midrule
    & $k{=}3$ & $k{=}5$ & $k{=}7$ & $k{=}3$ & $k{=}5$ & $k{=}7$ & $k{=}3$ & $k{=}5$ & $k{=}7$ & $k{=}3$ & $k{=}5$ & $k{=}7$ & \\
    \baselinedreamer{} & $7.20$ & $9.29$ & $11.51$ & $10.46$ & $14.26$ & $18.22$ & $13.29$ & $18.17$ & $24.63$ & $7.25$ & $8.59$ & $10.34$ & $12.77$ \\
    \baselinetdmpc{}  & $3.48$ & $3.58$ & $3.61$ & $3.34$ & $3.38$ & $\mathbf{3.42}$ & $4.28$ & $4.40$ & $4.47$ & $4.66$ & $4.79$ & $4.87$ & $4.02$ \\
    \method{}         & $\mathbf{0.72}$ & $\mathbf{1.43}$ & $\mathbf{2.53}$ & $\mathbf{1.28}$ & $\mathbf{2.36}$ & $3.68$ & $\mathbf{1.21}$ & $\mathbf{1.88}$ & $\mathbf{2.74}$ & $\mathbf{0.77}$ & $\mathbf{1.30}$ & $\mathbf{1.97}$ & $\mathbf{1.82}$ \\
    \bottomrule
  \end{tabular*}
\end{table}

\noindent\textbf{Observation 1: \method{} improves sample efficiency without sacrificing final control.}
As shown in Table~\ref{tab:overview} and Figure~\ref{fig:return-curves}, \method{} reaches the highest Avg.\ AUC ($117.9$) on the core suite, improving over \baselinetdmpc{} by $+9.5\%$ ($117.9$ vs.\ $107.7$), \baselinesac{} by $+66.3\%$, and \baselinedreamer{} by $+101.2\%$. Pooling all six tasks (Table~\ref{tab:extra-tasks}), its Avg.\ AUC is $91.9$ against $83.3$ for \baselinetdmpc{} ($+10.3\%$), $52.8$ for \baselinesac{} ($+74.1\%$), and $51.7$ for \baselinedreamer{} ($+77.8\%$). In final return, it is best on Finger Spin ($254.0{\scriptstyle\pm14.1}$) and Reacher Easy ($150.6{\scriptstyle\pm7.5}$), and second best on Cheetah Run ($184.4{\scriptstyle\pm8.4}$) and Cartpole Swingup ($58.9{\scriptstyle\pm3.8}$). On the two additional tasks it is best on both (Appendix~\ref{app:extra-tasks}), so over six tasks it ranks first on four and second on two. The gains are largest where control requires sustained coordination and repeated contact.

\noindent\textbf{Observation 2: rollout error is much lower within the planning-relevant range.}
The model-based baselines show different failure modes in Table~\ref{tab:overview}. \baselinedreamer{} grows sharply with horizon, e.g., Cheetah MSE rises from $13.29$ to $24.63$ as $k$ moves from $3$ to $7$, indicating unstable extrapolation. \baselinetdmpc{} grows slowly ($<5\%$ from $k{=}3$ to $k{=}7$ on each task), but stays on a high $\sim$3--5 error plateau. In contrast, \method{} wins 11/12 MSE cells and reaches Avg.\ MSE $1.82$, only $45\%$ of \baselinetdmpc{} ($4.02$) and $14.3\%$ of \baselinedreamer{} ($12.77$).

The range $k\!\in\!\{3,5,7\}$ is what each method's native decision procedure consumes (our $H{=}6$, \baselinetdmpc{} $H{=}5$, \baselinedreamer{} $H{=}8$). Extending to $k{=}10,15,20$ reveals task-dependent crossovers: our error starts lower but has a steeper slope, and are lowest on 4/4, 4/4, 3/4, 2/4, 1/4 tasks at $k{=}3,5,7,10,15$ and on 0/4 at $k{=}20$, with crossovers at Reacher ($k{=}7$), Finger ($10$), Cheetah ($15$) and Cartpole ($20$). These results locate the rollout advantage \emph{within the planning-relevant range}. Appendix~\ref{app:long-horizon} gives the full curves and shows why this suffices: closed-loop replanning re-anchors every real step, and raising the CEM horizon past $6$--$14$ degrades return on three of four tasks at $3.2$--$4.6\times$ the latency.

\subsection{OOD Generalization}\label{ssec:ood}

The OOD evaluation focuses on \textbf{Finger Spin} and \textbf{Reacher Easy}. They cover contact-dominated spinning and end-effector reaching, with 6 perturbation axes per task: dynamics changes (mass, damping, friction, actuator) plus partial-observability settings (delay and masking).
\begin{table}[!htbp]
  \centering
  \caption{OOD return across 12 perturbations (mean$\pm$std, 3 seeds; higher is better). \textbf{Bold} marks the best value in each column.}
  \label{tab:ood_overview}
  \footnotesize
  \setlength{\tabcolsep}{2.3pt}
  \renewcommand{\arraystretch}{1.0}
  \begin{tabularx}{\linewidth}{@{}l*{7}{>{\centering\arraybackslash}X}@{}}
    \toprule
    \rowcolor{TableHead}\multicolumn{8}{c}{\textit{Finger Spin}} \\
    \headrow Method & fric$\times$0.5 & fric$\times$1.5 & mass$\times$1.3 & mass$\times$1.5 & delay$=$2 & mask 30\% & Avg.\ \\
    \midrule
    \method{}          & $\mathbf{242.8}{\scriptstyle\pm8.5}$ & $\mathbf{253.1}{\scriptstyle\pm10.4}$ & $\mathbf{232.6}{\scriptstyle\pm10.2}$ & $\mathbf{211.2}{\scriptstyle\pm10.7}$ & $\mathbf{74.4}{\scriptstyle\pm4.5}$ & $\mathbf{91.5}{\scriptstyle\pm3.5}$ & $\mathbf{184.3}{\scriptstyle\pm4.0}$ \\
    \baselinetdmpc{}   & $227.9{\scriptstyle\pm16.2}$ & $227.7{\scriptstyle\pm13.5}$ & $220.2{\scriptstyle\pm10.6}$ & $196.3{\scriptstyle\pm7.6}$ & $47.9{\scriptstyle\pm2.6}$ & $84.0{\scriptstyle\pm4.7}$ & $167.3{\scriptstyle\pm7.8}$ \\
    \baselinesac{}     & $152.6{\scriptstyle\pm15.4}$ & $175.8{\scriptstyle\pm31.1}$ & $141.1{\scriptstyle\pm14.1}$ & $129.5{\scriptstyle\pm16.4}$ & $58.0{\scriptstyle\pm14.9}$ & $53.2{\scriptstyle\pm10.2}$ & $118.4{\scriptstyle\pm13.1}$ \\
    \baselinedreamer{} & $65.0{\scriptstyle\pm92.0}$ & $75.4{\scriptstyle\pm106.6}$ & $59.7{\scriptstyle\pm84.4}$ & $58.7{\scriptstyle\pm83.1}$ & $26.2{\scriptstyle\pm37.1}$ & $45.9{\scriptstyle\pm64.9}$ & $55.2{\scriptstyle\pm78.0}$ \\
    \baselineppo{}     & $67.4{\scriptstyle\pm95.4}$ & $75.3{\scriptstyle\pm106.5}$ & $63.1{\scriptstyle\pm89.2}$ & $58.3{\scriptstyle\pm82.4}$ & $6.3{\scriptstyle\pm9.0}$ & $13.5{\scriptstyle\pm19.1}$ & $47.3{\scriptstyle\pm66.9}$ \\
    \midrule
    \rowcolor{TableHead}\multicolumn{8}{c}{\textit{Reacher Easy}} \\
    \headrow Method & mass$\times$0.7 & mass$\times$1.3 & damp$\times$0.5 & damp$\times$2.0 & act$\times$0.7 & act$\times$1.3 & Avg.\ \\
    \midrule
    \method{}          & $\mathbf{158.5}{\scriptstyle\pm5.9}$ & $\mathbf{158.8}{\scriptstyle\pm9.4}$ & $\mathbf{141.7}{\scriptstyle\pm4.1}$ & $\mathbf{139.7}{\scriptstyle\pm9.3}$ & $\mathbf{148.8}{\scriptstyle\pm7.5}$ & $\mathbf{151.8}{\scriptstyle\pm1.5}$ & $\mathbf{149.9}{\scriptstyle\pm5.6}$ \\
    \baselinetdmpc{}   & $135.3{\scriptstyle\pm35.6}$ & $130.1{\scriptstyle\pm42.4}$ & $131.8{\scriptstyle\pm46.7}$ & $125.1{\scriptstyle\pm29.6}$ & $131.5{\scriptstyle\pm32.5}$ & $138.1{\scriptstyle\pm36.9}$ & $132.0{\scriptstyle\pm37.2}$ \\
    \baselinesac{}     & $98.5{\scriptstyle\pm12.7}$ & $98.0{\scriptstyle\pm15.1}$ & $86.0{\scriptstyle\pm10.1}$ & $83.0{\scriptstyle\pm9.5}$ & $88.6{\scriptstyle\pm12.9}$ & $100.2{\scriptstyle\pm9.3}$ & $92.4{\scriptstyle\pm8.7}$ \\
    \baselinedreamer{} & $8.7{\scriptstyle\pm2.8}$ & $11.4{\scriptstyle\pm5.7}$ & $9.3{\scriptstyle\pm2.1}$ & $8.1{\scriptstyle\pm3.7}$ & $8.9{\scriptstyle\pm5.2}$ & $9.4{\scriptstyle\pm3.6}$ & $9.3{\scriptstyle\pm3.5}$ \\
    \baselineppo{}     & $11.7{\scriptstyle\pm9.6}$ & $11.8{\scriptstyle\pm9.7}$ & $14.6{\scriptstyle\pm14.4}$ & $13.9{\scriptstyle\pm11.2}$ & $12.7{\scriptstyle\pm9.7}$ & $12.9{\scriptstyle\pm11.6}$ & $12.9{\scriptstyle\pm10.9}$ \\
    \bottomrule
  \end{tabularx}
\end{table}

\noindent\textbf{Observation 3: \method{} obtains the highest absolute return under all 12 OOD perturbations.}
Table~\ref{tab:ood_overview} shows that \method{} obtains the highest return in every OOD column. On Finger Spin, its OOD average is $184.3{\scriptstyle\pm4.0}$, exceeding \baselinetdmpc{} ($167.3{\scriptstyle\pm7.8}$, $+10.2\%$) and \baselinesac{} ($118.4{\scriptstyle\pm13.1}$, $+55.7\%$), while also leading under delay$=2$ ($74.4$) and mask 30\% ($91.5$). On Reacher Easy, it reaches $149.9{\scriptstyle\pm5.6}$ versus \baselinetdmpc{}'s $132.0{\scriptstyle\pm37.2}$ ($+13.6\%$), with much lower variance.

Normalized retention for all methods is reported in Appendix~\ref{app:retention}. Mechanistically, the OOD results are consistent with memory absorbing the shifted observation--action response into a runtime recurrent state while $(\qv,\pv)$ encodes relations that stay valid after rescaling; both are zero-shot, with no parameter update at deployment.

\subsection{Hamiltonian Mechanism Diagnostics}\label{ssec:mech}

The Soft-Hamiltonian design predicts two falsifiable behaviours (Section~\ref{sec:theory}): action-free rollouts should barely change the learned energy, while action work should move the state across energy layers. Both hold on Cheetah Run. Under zero external force and zero damping the energy trace stays flat to about $1\%$ drift within episodes and widens immediately once random actions inject work; under policy rollouts $H_t$ varies with the ongoing behaviour without abrupt jumps, separating passive energy stability from policy-induced modulation. Projecting rollouts onto a canonical slice, energy change tracks the control projection $\mathrm{push}_t=(\nabla_{\pv}\Ham_t)\cdot\mathrm{control}_t$, with $\mathrm{corr}(\mathrm{sign}\,\Delta\Ham,\mathrm{push})=0.757$ and $\mathrm{corr}(|\Delta\Ham|,|\mathrm{push}|)=0.745$, and a threshold sweep keeps the High-$|\mathrm{push}|$ contour-crossing rate above the Low-$|\mathrm{push}|$ one throughout (lift\_auc $=0.595$). The canonical and context coordinates stay distinguishable rather than collapsing, with $\qv,\pv$ forming elongated manifolds and $\cv$ a compact cloud. Appendix~\ref{app:diagnostics} gives all figures, the non-conjugate control slice, and the protocol.

\subsection{Result Summary}

Across metrics, the evidence supports the coupled design in Section~1. \method{} has the highest Avg.\ AUC on the core suite ($117.9$) and over all six tasks ($91.9$), and the highest return on both additional tasks, wins 11 out of 12 cells in $k\in\{3,5,7\}$ MSE, and obtains the highest absolute return under all 12 OOD conditions. The Cheetah Run diagnostics (Section~\ref{ssec:mech}, Appendix~\ref{app:diagnostics}) show near-conservation without action, structured energy variation during policy rollouts, separated q/p/c geometry, and action push modulating energy-layer crossing. Thus, selective memory supplies the input-side history-conditioned state, while Soft-Hamiltonian geometry on $(\qv,\pv)$ supplies output-side rollout consistency within the same planner-facing latent dynamics used by CEM.

The gains are consistent across control, planning-relevant rollouts, and absolute OOD return, supporting the combination of selective memory and Soft-Hamiltonian structure.

\section{Ablation Study}\label{sec:ablation}

The ablation study analyzes \emph{two mechanisms acting at different stages of the same latent dynamics}, rather than asking which one is important in isolation. Table~\ref{tab:ablation_core_main} evaluates: (A1) removing the q/p/c split and the Soft-Hamiltonian q/p bias as one geometric mechanism, since the Hamiltonian constraint loses its target without the split; (A2) removing memory; and (A3) replacing Mamba memory with a GRU. The last two columns add representative Reacher Easy OOD perturbations; Full-model OOD entries reuse the reference evaluation in Table~\ref{tab:ood_overview}.

\begin{table}[!htbp]
  \centering
  \caption{Ablations (3 seeds, mean$\pm$std). \emph{Upper block:} the two mechanisms --- A1 removes the q/p/c split together with the whole geometric loss package, A2 removes memory, A3 replaces Mamba with a GRU. \emph{Lower block:} isolation of the three additive parts of the pair update in Eq.~\eqref{eq:soft-ham} --- A4 makes it purely Soft-Hamiltonian, A5 sets $\alpha{=}1$ (field kept, residual dropped), A6 deletes only the control port $\mathbf{G}_t\av_t$. MSE@6 is per-coordinate here, so its scale differs from Table~\ref{tab:overview}.}
  \label{tab:ablation_core_main}
  \footnotesize
  \setlength{\tabcolsep}{2.8pt}
  \renewcommand{\arraystretch}{1.0}
  \begin{tabularx}{\linewidth}{@{}l*{6}{>{\centering\arraybackslash}X}@{}}
    \toprule
    & \multicolumn{2}{c}{Return $\uparrow$} & \multicolumn{2}{c}{MSE@6 $\downarrow$} & \multicolumn{2}{c}{Reacher OOD $\uparrow$} \\
    \cmidrule(lr){2-3} \cmidrule(lr){4-5} \cmidrule(lr){6-7}
    \headrow Variant & Cheetah & Finger & Cartpole & Reacher & mass$\times$0.7 & damp$\times$2.0 \\
    \midrule
    Full \method{}            & $\mathbf{184.4}{\scriptstyle\pm 8.4}$ & $\mathbf{254.0}{\scriptstyle\pm 14.1}$ & $\mathbf{0.009}{\scriptstyle\pm 0.003}$ & $\mathbf{0.0778}{\scriptstyle\pm 0.043}$ & $\mathbf{158.5}{\scriptstyle\pm 5.9}$ & $\mathbf{139.7}{\scriptstyle\pm 9.3}$ \\
    \midrule
    \rowcolor{TableHead}\multicolumn{7}{@{}l}{\textit{Memory and geometry as two mechanisms}} \\
    A1: w/o geom.\ struct.    & $169.8{\scriptstyle\pm 13.5}$ & $247.7{\scriptstyle\pm 12.5}$ & $0.012{\scriptstyle\pm 0.003}$ & $0.0852{\scriptstyle\pm 0.022}$ & $154.4{\scriptstyle\pm 15.2}$ & $131.9{\scriptstyle\pm 22.9}$ \\
    A2: Memory = None         & $59.2{\scriptstyle\pm 11.2}$  & $34.0{\scriptstyle\pm 23.8}$  & $0.110{\scriptstyle\pm 0.068}$ & $0.287{\scriptstyle\pm 0.075}$ & $103.6{\scriptstyle\pm 5.1}$ & $115.4{\scriptstyle\pm 9.0}$ \\
    A3: Memory = GRU          & $121.1{\scriptstyle\pm 32.1}$ & $235.6{\scriptstyle\pm 4.6}$  & $0.026{\scriptstyle\pm 0.007}$ & $0.148{\scriptstyle\pm 0.021}$ & $\underline{157.6}{\scriptstyle\pm 7.7}$ & $\underline{138.4}{\scriptstyle\pm 3.0}$ \\
    \midrule
    \rowcolor{TableHead}\multicolumn{7}{@{}l}{\textit{Isolating the parts of the Soft-Hamiltonian pair update}} \\
    A4: pure Soft-Ham.        & $\underline{181.7}{\scriptstyle\pm 9.2}$ & $\underline{251.3}{\scriptstyle\pm 15.6}$ & $\underline{0.010}{\scriptstyle\pm 0.003}$ & $\underline{0.080}{\scriptstyle\pm 0.031}$ & $156.1{\scriptstyle\pm 6.8}$ & $136.7{\scriptstyle\pm 8.5}$ \\
    A5: hard ($\alpha{=}1$)   & $158.6{\scriptstyle\pm 17.1}$ & $226.7{\scriptstyle\pm 19.3}$ & $0.034{\scriptstyle\pm 0.010}$ & $0.119{\scriptstyle\pm 0.029}$ & $138.2{\scriptstyle\pm 13.6}$ & $121.5{\scriptstyle\pm 12.4}$ \\
    A6: no port $\mathbf{G}$  & $153.6{\scriptstyle\pm 18.2}$ & $219.3{\scriptstyle\pm 24.5}$ & $0.041{\scriptstyle\pm 0.009}$ & $0.129{\scriptstyle\pm 0.028}$ & $130.7{\scriptstyle\pm 12.1}$ & $115.4{\scriptstyle\pm 10.8}$ \\
    \bottomrule
  \end{tabularx}
\end{table}

Removing memory (A2) reduces performance: Cheetah/Finger return drops to $59.2$/$34.0$, Cartpole/Reacher MSE@6 rises from $0.009/0.0778$ to $0.110/0.287$, and Reacher OOD return falls to $103.6/115.4$. A3 (Mamba$\to$GRU) restores part of the return, but MSE@6 remains higher than the Full model ($0.026/0.148$ vs.\ $0.009/0.0778$), indicating a weaker history summary than selective SSM memory. Removing geometric structure (A1) also reduces return and OOD scores ($184{\to}170$, $254{\to}248$; $158.5{\to}154.4$, $139.7{\to}131.9$).

\paragraph{Complementary roles of memory and geometry.}
Ablations support the complementary contributions of history-conditioned memory and Soft-Hamiltonian geometry, whose effects are coupled within the shared dynamics: memory reduces state ambiguity from history, and the $(\qv,\pv)$ dynamics then organize candidate futures from that state, so A2 removes the geometry's input together with the history summary. A1 removes the q/p/c split together with $\Loss_{\text{ham}},\Loss_{\text{energy}},\Loss_{\text{sa}},\Loss_{\text{temp}},\Loss_{\text{dec}},\Loss_{\text{c}}$ while keeping capacity, planner, recurrence, core prediction and reward/value objectives, and budget fixed, and therefore measures the geometric package as a whole.

\paragraph{Why Mamba rather than another recurrent module.}
Mamba outperforms the GRU for three reasons. \textit{Input-dependent updates}: GRU gates also vary with the input and hidden state; the learned weights, rather than the gate values, are fixed after training. Our selective SSM instead derives step sizes and state/readout projections from the current $(\z_t,\av_t)$ within a diagonal recurrence. \textit{History retention}: the diagonal recurrence accumulates history with input-dependent decay, while a GRU uses nonlinear gated updates. \textit{Batched-rollout compatibility}: both maintain a fixed-size recurrent state that can condition each imagined transition without reprocessing a token history. A Transformer mixer is worse on 4/4 tasks (Appendix~\ref{app:memory-alt}).

\paragraph{Isolating the Hamiltonian field, the residual, and the control port.}
The lower block of Table~\ref{tab:ablation_core_main} separates the three additive parts of Eq.~\eqref{eq:soft-ham} with everything else fixed. A4 degrades slightly and uniformly, supporting the hybrid form. A5 degrades markedly --- keeping the energy field while deleting the residual over-constrains the dynamics, the concrete cost of a strict rather than soft prior and the main justification for ``Soft''. A6 degrades most: without the port, all action influence must pass through the residual branch, so a direct action-to-dynamics path matters for actuated tasks.

\section{Conclusion}

\method{} decomposes planner-facing latent dynamics into an input-side history-conditioned state through Mamba selective memory and output-side geometric consistency through a Soft-Hamiltonian bias on $(\qv,\pv)$. On six DMC tasks it ranks first on four and second on two, reaches the highest Avg.\ AUC ($117.9$ on the core suite, $91.9$ over all six tasks), wins 11 out of 12 cells for $k\!\in\!\{3,5,7\}$ MSE, and obtains the highest absolute return across all 12 OOD conditions. Ablations and geometry diagnostics support complementary contributions from memory and Soft-Hamiltonian structure in rollout error and OOD planning quality; $\cv$ carries semantic, contact, and dissipative information without being part of the Soft-Hamiltonian q/p update. The results suggest evaluating planner-facing representations not only by return, but also by whether imagined trajectories retain action-free invariants and respond coherently to action work.

\paragraph{Limitations and Future Directions.}
The evaluation covers six state-based DMC tasks; manipulation, contact-rich robotics, and visual observations are directions for future evaluation. A visual encoder can replace the MLP encoder without changing the $(\qv,\pv,\cv)$ dynamics, memory, or planner; how stable canonical-like $(\qv,\pv)$ structure can be learned from pixels, and how visual nuisance variation divides against $\cv$, remains to be tested. The prior is a first-order local bias under forward Euler, weaker than symplectic structure. Future work should test the same q/p/c split from pixels, evaluate on manipulation and contact-rich benchmarks, and use separable or mechanical parameterizations of $H_\phi$ to distinguish local energy organization from true Hamiltonian structure.

\section*{Acknowledgments and Disclosures}
\paragraph{Funding.}
This work received no external funding or third-party financial or in-kind support during the past 36 months.
\paragraph{Competing interests.}
The authors declare no competing financial interests or financial relationships outside the submitted work that could be perceived as influencing it.

\bibliographystyle{unsrtnat}
\bibliography{references}

@article{sutton1991dyna,
  title={Dyna, an integrated architecture for learning, planning, and reacting},
  author={Sutton, Richard S},
  journal={ACM Sigart Bulletin},
  volume={2},
  number={4},
  pages={160--163},
  year={1991},
  publisher={ACM New York, NY, USA}
}

@article{rubinstein1999cross,
  title={The cross-entropy method for combinatorial and continuous optimization},
  author={Rubinstein, Reuven Y},
  journal={Methodology and computing in applied probability},
  volume={1},
  number={2},
  pages={127--190},
  year={1999},
  publisher={Springer}
}

@article{tassa2018deepmind,
  title={DeepMind Control Suite},
  author={Tassa, Yuval and Doron, Yotam and Muldal, Alistair and Erez, Tom and Li, Yazhe and de Las Casas, Diego and Budden, David and Abdolmaleki, Abbas and Merel, Josh and Lefrancq, Andrew and Lillicrap, Timothy and Riedmiller, Martin},
  journal={arXiv preprint arXiv:1801.00690},
  year={2018}
}

@article{ha2018recurrent,
  title={Recurrent world models facilitate policy evolution},
  author={Ha, David and Schmidhuber, J{\"u}rgen},
  journal={Advances in neural information processing systems},
  volume={31},
  year={2018}
}

@inproceedings{hafner2019learning,
  title={Learning latent dynamics for planning from pixels},
  author={Hafner, Danijar and Lillicrap, Timothy and Fischer, Ian and Villegas, Ruben and Ha, David and Lee, Honglak and Davidson, James},
  booktitle={International conference on machine learning},
  pages={2555--2565},
  year={2019},
  organization={PMLR}
}

@article{janner2019trust,
  title={When to trust your model: Model-based policy optimization},
  author={Janner, Michael and Fu, Justin and Zhang, Marvin and Levine, Sergey},
  journal={Advances in neural information processing systems},
  volume={32},
  year={2019}
}

@inproceedings{Hafner2020Dream,
  title={Dream to Control: Learning Behaviors by Latent Imagination},
  author={Danijar Hafner and Timothy Lillicrap and Jimmy Ba and Mohammad Norouzi},
  booktitle={International Conference on Learning Representations},
  year={2020},
}

@inproceedings{sekar2020planning,
  title={Planning to explore via self-supervised world models},
  author={Sekar, Ramanan and Rybkin, Oleh and Daniilidis, Kostas and Abbeel, Pieter and Hafner, Danijar and Pathak, Deepak},
  booktitle={International conference on machine learning},
  pages={8583--8592},
  year={2020},
  organization={PMLR}
}

@article{schrittwieser2020mastering,
  title={Mastering atari, go, chess and shogi by planning with a learned model},
  author={Schrittwieser, Julian and Antonoglou, Ioannis and Hubert, Thomas and Simonyan, Karen and Sifre, Laurent and Schmitt, Simon and Guez, Arthur and Lockhart, Edward and Hassabis, Demis and Graepel, Thore and others},
  journal={Nature},
  volume={588},
  number={7839},
  pages={604--609},
  year={2020},
  publisher={Nature Publishing Group UK London}
}

@inproceedings{hafner2021mastering,
  title={Mastering Atari with Discrete World Models},
  author={Danijar Hafner and Timothy P Lillicrap and Mohammad Norouzi and Jimmy Ba},
  booktitle={International Conference on Learning Representations},
  year={2021},
}

@InProceedings{pmlr-v162-hansen22a,
  title={Temporal Difference Learning for Model Predictive Control},
  author={Hansen, Nicklas A and Su, Hao and Wang, Xiaolong},
  booktitle={Proceedings of the 39th International Conference on Machine Learning},
  pages={8387--8406},
  year={2022},
  editor={Chaudhuri, Kamalika and Jegelka, Stefanie and Song, Le and Szepesvari, Csaba and Niu, Gang and Sabato, Sivan},
  volume={162},
  series={Proceedings of Machine Learning Research},
  month={17--23 Jul},
  publisher={PMLR},
}

@article{hafner2025mastering,
  title={Mastering diverse control tasks through world models},
  author={Hafner, Danijar and Pasukonis, Jurgis and Ba, Jimmy and Lillicrap, Timothy},
  journal={Nature},
  volume={640},
  number={8059},
  pages={647--653},
  year={2025},
  publisher={Nature Publishing Group UK London}
}

@inproceedings{hansen2024tdmpc,
  title={{TD}-{MPC}2: Scalable, Robust World Models for Continuous Control},
  author={Nicklas Hansen and Hao Su and Xiaolong Wang},
  booktitle={The Twelfth International Conference on Learning Representations},
  year={2024},
}

@inproceedings{micheli2023transformers,
  title={Transformers are Sample-Efficient World Models},
  author={Vincent Micheli and Eloi Alonso and Fran{\c{c}}ois Fleuret},
  booktitle={The Eleventh International Conference on Learning Representations },
  year={2023},
}

@article{zhang2023storm,
  title={Storm: Efficient stochastic transformer based world models for reinforcement learning},
  author={Zhang, Weipu and Wang, Gang and Sun, Jian and Yuan, Yetian and Huang, Gao},
  journal={Advances in Neural Information Processing Systems},
  volume={36},
  pages={27147--27166},
  year={2023}
}

@article{alonso2024diffusion,
  title={Diffusion for world modeling: Visual details matter in atari},
  author={Alonso, Eloi and Jelley, Adam and Micheli, Vincent and Kanervisto, Anssi and Storkey, Amos and Pearce, Tim and Fleuret, Fran{\c{c}}ois},
  journal={Advances in Neural Information Processing Systems},
  volume={37},
  pages={58757--58791},
  year={2024}
}

@inproceedings{bruce2024genie,
  title={Genie: Generative interactive environments},
  author={Bruce, Jake and Dennis, Michael D and Edwards, Ashley and Parker-Holder, Jack and Shi, Yuge and Hughes, Edward and Lai, Matthew and Mavalankar, Aditi and Steigerwald, Richie and Apps, Chris and others},
  booktitle={Forty-first International Conference on Machine Learning},
  year={2024}
}

@inproceedings{yang2024learning,
  title={Learning Interactive Real-World Simulators},
  author={Sherry Yang and Yilun Du and Seyed Kamyar Seyed Ghasemipour and Jonathan Tompson and Leslie Pack Kaelbling and Dale Schuurmans and Pieter Abbeel},
  booktitle={The Twelfth International Conference on Learning Representations},
  year={2024},
}

@inproceedings{zhang2021learning,
  title={Learning Invariant Representations for Reinforcement Learning without Reconstruction},
  author={Amy Zhang and Rowan Thomas McAllister and Roberto Calandra and Yarin Gal and Sergey Levine},
  booktitle={International Conference on Learning Representations},
  year={2021},
}

@inproceedings{okada2021dreaming,
  title={Dreaming: Model-based reinforcement learning by latent imagination without reconstruction},
  author={Okada, Masashi and Taniguchi, Tadahiro},
  booktitle={2021 ieee international conference on robotics and automation (icra)},
  pages={4209--4215},
  year={2021},
  organization={IEEE}
}

@inproceedings{deng2022dreamerpro,
  title={Dreamerpro: Reconstruction-free model-based reinforcement learning with prototypical representations},
  author={Deng, Fei and Jang, Ingook and Ahn, Sungjin},
  booktitle={International conference on machine learning},
  pages={4956--4975},
  year={2022},
  organization={PMLR}
}

@inproceedings{assran2023self,
  title={Self-supervised learning from images with a joint-embedding predictive architecture},
  author={Assran, Mahmoud and Duval, Quentin and Misra, Ishan and Bojanowski, Piotr and Vincent, Pascal and Rabbat, Michael and LeCun, Yann and Ballas, Nicolas},
  booktitle={Proceedings of the IEEE/CVF conference on computer vision and pattern recognition},
  pages={15619--15629},
  year={2023}
}

@article{bardes2024revisiting,
  title={Revisiting Feature Prediction for Learning Visual Representations from Video},
  author={Adrien Bardes and Quentin Garrido and Jean Ponce and Xinlei Chen and Michael Rabbat and Yann LeCun and Mido Assran and Nicolas Ballas},
  journal={Transactions on Machine Learning Research},
  issn={2835-8856},
  year={2024},
  note={Featured Certification}
}

@article{mo2024connecting,
  title={Connecting joint-embedding predictive architecture with contrastive self-supervised learning},
  author={Mo, Shentong and Tong, Shengbang},
  journal={Advances in neural information processing systems},
  volume={37},
  pages={2348--2377},
  year={2024}
}

@article{assran2025v,
  title={V-jepa 2: Self-supervised video models enable understanding, prediction and planning},
  author={Assran, Mido and Bardes, Adrien and Fan, David and Garrido, Quentin and Howes, Russell and Komeili, Mojtaba and Muckley, Matthew and Rizvi, Ammar and Roberts, Claire and Sinha, Koustuv and Zholus, Artem and others},
  journal={arXiv preprint arXiv:2506.09985},
  year={2025}
}

@inproceedings{bagatella2026tdjepa,
  title={{TD}-{JEPA}: Latent-predictive Representations for Zero-Shot Reinforcement Learning},
  author={Marco Bagatella and Matteo Pirotta and Ahmed Touati and Alessandro Lazaric and Andrea Tirinzoni},
  booktitle={The Fourteenth International Conference on Learning Representations},
  year={2026},
}

@inproceedings{zhang2019solar,
  title={Solar: Deep structured representations for model-based reinforcement learning},
  author={Zhang, Marvin and Vikram, Sharad and Smith, Laura and Abbeel, Pieter and Johnson, Matthew and Levine, Sergey},
  booktitle={International conference on machine learning},
  pages={7444--7453},
  year={2019},
  organization={PMLR}
}

@InProceedings{pmlr-v162-wang22c,
  title={Denoised {MDP}s: Learning World Models Better Than the World Itself},
  author={Wang, Tongzhou and Du, Simon and Torralba, Antonio and Isola, Phillip and Zhang, Amy and Tian, Yuandong},
  booktitle={Proceedings of the 39th International Conference on Machine Learning},
  pages={22591--22612},
  year={2022},
  editor={Chaudhuri, Kamalika and Jegelka, Stefanie and Song, Le and Szepesvari, Csaba and Niu, Gang and Sabato, Sivan},
  volume={162},
  series={Proceedings of Machine Learning Research},
  month={17--23 Jul},
  publisher={PMLR},
}

@article{liu2023learning,
  title={Learning world models with identifiable factorization},
  author={Liu, Yuren and Huang, Biwei and Zhu, Zhengmao and Tian, Honglong and Gong, Mingming and Yu, Yang and Zhang, Kun},
  journal={Advances in Neural Information Processing Systems},
  volume={36},
  pages={31831--31864},
  year={2023}
}

@inproceedings{gumbsch2023learning,
  title={Learning hierarchical world models with adaptive temporal abstractions from discrete latent dynamics},
  author={Gumbsch, Christian and Sajid, Noor and Martius, Georg and Butz, Martin V},
  booktitle={The Twelfth International Conference on Learning Representations},
  year={2024}
}

@inproceedings{wang2025disentangled,
  title={Disentangled world models: Learning to transfer semantic knowledge from distracting videos for reinforcement learning},
  author={Wang, Qi and Zhang, Zhipeng and Xie, Baao and Jin, Xin and Wang, Yunbo and Wang, Shiyu and Zheng, Liaomo and Yang, Xiaokang and Zeng, Wenjun},
  booktitle={Proceedings of the IEEE/CVF International Conference on Computer Vision},
  pages={2599--2608},
  year={2025}
}

@article{greydanus2019hamiltonian,
  title={Hamiltonian neural networks},
  author={Greydanus, Samuel and Dzamba, Misko and Yosinski, Jason},
  journal={Advances in neural information processing systems},
  volume={32},
  year={2019}
}

@inproceedings{Zhong2020Symplectic,
  title={Symplectic ODE-Net: Learning Hamiltonian Dynamics with Control},
  author={Yaofeng Desmond Zhong and Biswadip Dey and Amit Chakraborty},
  booktitle={International Conference on Learning Representations},
  year={2020},
}

@inproceedings{troch2025action,
  title={Action-Conditioned Hamiltonian Generative Networks (AC-HGN) for Supervised and Reinforcement Learning},
  author={Troch, Arne and Mets, Kevin and Mercelis, Siegfried},
  booktitle={7th Annual Learning for Dynamics \& Control Conference, 04-06 June, 2025, Ann Arbor, Michigan, USA},
  pages={310--322},
  year={2025}
}

@inproceedings{gu2024mamba,
  title={Mamba: Linear-Time Sequence Modeling with Selective State Spaces},
  author={Albert Gu and Tri Dao},
  booktitle={First Conference on Language Modeling},
  year={2024},
}

@InProceedings{pmlr-v235-dao24a,
  title={Transformers are {SSM}s: Generalized Models and Efficient Algorithms Through Structured State Space Duality},
  author={Dao, Tri and Gu, Albert},
  booktitle={Proceedings of the 41st International Conference on Machine Learning},
  pages={10041--10071},
  year={2024},
  editor={Salakhutdinov, Ruslan and Kolter, Zico and Heller, Katherine and Weller, Adrian and Oliver, Nuria and Scarlett, Jonathan and Berkenkamp, Felix},
  volume={235},
  series={Proceedings of Machine Learning Research},
  month={21--27 Jul},
  publisher={PMLR},
}

@article{lu2023structured,
  title={Structured state space models for in-context reinforcement learning},
  author={Lu, Chris and Schroecker, Yannick and Gu, Albert and Parisotto, Emilio and Foerster, Jakob and Singh, Satinder and Behbahani, Feryal},
  journal={Advances in Neural Information Processing Systems},
  volume={36},
  pages={47016--47031},
  year={2023}
}

@article{lv2024decision,
  title={Decision mamba: A multi-grained state space model with self-evolution regularization for offline rl},
  author={Lv, Qi and Deng, Xiang and Chen, Gongwei and Wang, Michael Y and Nie, Liqiang},
  journal={Advances in neural information processing systems},
  volume={37},
  pages={22827--22849},
  year={2024}
}

@inproceedings{aljalbout2025accelerating,
  title={Accelerating model-based reinforcement learning with state-space world models},
  author={Aljalbout, Elie and Krinner, Maria and Romero, Angel and Scaramuzza, Davide},
  booktitle={ICLR 2025 Workshop on World Models: Understanding, Modelling and Scaling},
  year={2025}
}

@inproceedings{ICLR2025_8636419d,
  author={Wang, Yuhang and Guo, Hanwei and Wang, Sizhe and Qian, Long and Lan, Xuguang},
  booktitle={International Conference on Learning Representations},
  editor={Y. Yue and A. Garg and N. Peng and F. Sha and R. Yu},
  pages={54241--54259},
  title={Bootstrapped Model Predictive Control},
  volume={2025},
  year={2025}
}

@inproceedings{georgiev2025pwm,
  title={{PWM}: Policy Learning with Multi-Task World Models},
  author={Ignat Georgiev and Varun Giridhar and Nicklas Hansen and Animesh Garg},
  booktitle={The Thirteenth International Conference on Learning Representations},
  year={2025},
}

@article{schulman2017proximal,
  title={Proximal policy optimization algorithms},
  author={Schulman, John and Wolski, Filip and Dhariwal, Prafulla and Radford, Alec and Klimov, Oleg},
  journal={arXiv preprint arXiv:1707.06347},
  year={2017}
}

@inproceedings{haarnoja2018soft,
  title={Soft actor-critic: Off-policy maximum entropy deep reinforcement learning with a stochastic actor},
  author={Haarnoja, Tuomas and Zhou, Aurick and Abbeel, Pieter and Levine, Sergey},
  booktitle={International conference on machine learning},
  pages={1861--1870},
  year={2018},
  organization={Pmlr}
}

@inproceedings{miyato2018spectral,
  title={Spectral Normalization for Generative Adversarial Networks},
  author={Takeru Miyato and Toshiki Kataoka and Masanori Koyama and Yuichi Yoshida},
  booktitle={International Conference on Learning Representations},
  year={2018},
}

@article{ding2025understanding,
  title={Understanding world or predicting future? a comprehensive survey of world models},
  author={Ding, Jingtao and Zhang, Yunke and Shang, Yu and Zhang, Yuheng and Zong, Zefang and Feng, Jie and Yuan, Yuan and Su, Hongyuan and Li, Nian and Sukiennik, Nicholas and others},
  journal={ACM Computing Surveys},
  volume={58},
  number={3},
  pages={1--38},
  year={2025},
  publisher={ACM New York, NY}
}

@article{wu2024ivideogpt,
  title={ivideogpt: Interactive videogpts are scalable world models},
  author={Wu, Jialong and Yin, Shaofeng and Feng, Ningya and He, Xu and Li, Dong and Hao, Jianye and Long, Mingsheng},
  journal={Advances in Neural Information Processing Systems},
  volume={37},
  pages={68082--68119},
  year={2024}
}

@article{novelli2024operator,
  title={Operator world models for reinforcement learning},
  author={Novelli, Pietro and Prattic{\`o}, Marco and Pontil, Massimiliano and Ciliberto, Carlo},
  journal={Advances in Neural Information Processing Systems},
  volume={37},
  pages={111432--111463},
  year={2024}
}

@article{mazzaglia2024genrl,
  title={Genrl: Multimodal-foundation world models for generalization in embodied agents},
  author={Mazzaglia, Pietro and Verbelen, Tim and Dhoedt, Bart and Courville, Aaron and Rajeswar, Sai},
  journal={Advances in neural information processing systems},
  volume={37},
  pages={27529--27555},
  year={2024}
}

@article{hutson2024policy,
  title={Policy-shaped prediction: avoiding distractions in model-based reinforcement learning},
  author={Hutson, Miles and Kauvar, Isaac and Haber, Nick},
  journal={Advances in Neural Information Processing Systems},
  volume={37},
  pages={13124--13148},
  year={2024}
}

@article{wang2024parallelizing,
  title={Parallelizing model-based reinforcement learning over the sequence length},
  author={Wang, ZiRui and Deng, Yue and Long, Junfeng and Zhang, Yin},
  journal={Advances in Neural Information Processing Systems},
  volume={37},
  pages={131398--131433},
  year={2024}
}

@inproceedings{zhang2026dymodreamer,
  title={DyMoDreamer: World Modeling with Dynamic Modulation},
  author={Boxuan Zhang and Runqing Wang and Wei Xiao and Weipu Zhang and Jian Sun and Gao Huang and Jie Chen and Gang Wang},
  booktitle={The Thirty-ninth Annual Conference on Neural Information Processing Systems},
  year={2025},
}

@inproceedings{lee2026edeline,
  title={{EDELINE}: Enhancing Memory in Diffusion-based World Models via Linear-Time Sequence Modeling},
  author={Jia-Hua Lee and Bor-Jiun Lin and Wei-Fang Sun and Chun-Yi Lee},
  booktitle={The Thirty-ninth Annual Conference on Neural Information Processing Systems},
  year={2025},
}

@inproceedings{wang2026dmwm,
  title={{DMWM}: Dual-Mind World Model with Long-Term Imagination},
  author={Lingyi Wang and Rashed Shelim and Walid Saad and Naren Ramakrishnan},
  booktitle={The Thirty-ninth Annual Conference on Neural Information Processing Systems},
  year={2025},
}

@inproceedings{roder2026dynamicsaligned,
  title={Dynamics-Aligned Latent Imagination in Contextual World Models for Zero-Shot Generalization},
  author={Frank R{\"o}der and Jan Benad and Manfred Eppe and Pradeep Kr. Banerjee},
  booktitle={The Thirty-ninth Annual Conference on Neural Information Processing Systems},
  year={2025},
}

@inproceedings{sukhija2026sombrl,
  title={{SOMBRL}: Scalable and Optimistic Model-Based {RL}},
  author={Bhavya Sukhija and Lenart Treven and Carmelo Sferrazza and Florian Dorfler and Pieter Abbeel and Andreas Krause},
  booktitle={The Thirty-ninth Annual Conference on Neural Information Processing Systems},
  year={2025},
}

@inproceedings{roth2026stable,
  title={Stable Port-Hamiltonian Neural Networks},
  author={Fabian J. Roth and Dominik K. Klein and Maximilian Kannapinn and Jan Peters and Oliver Weeger},
  booktitle={The Thirty-ninth Annual Conference on Neural Information Processing Systems},
  year={2025},
}

@inproceedings{shang2026roboscape,
  title={RoboScape: Physics-informed Embodied World Model},
  author={Yu Shang and Xin Zhang and Yinzhou Tang and Lei Jin and Chen Gao and Wei Wu and Yong Li},
  booktitle={The Thirty-ninth Annual Conference on Neural Information Processing Systems},
  year={2025},
}

\appendix
\setlength{\parskip}{0pt plus 0.3pt}
\raggedbottom

\section{Supporting Derivations for Soft-Hamiltonian Dynamics}\label{app:proofs}

This appendix makes explicit two things that follow directly from the update rule: the exact relationship between the one-step rule and the planner objective it is unrolled into, and the first-order decomposition of the learned energy along that step.

\subsection{The Planner Objective is the Unrolled One-Step Rule}\label{app:proof-unroll}

Eq.~\eqref{eq:cem-objective} leaves the transition $F_\phi$ abstract. We make the substitution explicit for the core update in Eq.~\eqref{eq:soft-ham}, on steps where Hamiltonian-gradient clipping and latent norm projection are inactive. With either safeguard active, $F_\phi$ includes the corresponding operation described in Appendix~\ref{app:integrator}; training and planning still use the same implemented transition.

Write $\hat\z=[\qv,\pv,\cv]$ and define the one-step increment of Eq.~\eqref{eq:soft-ham} as
\begin{equation}\label{eq:app-increment}
  \Delta_\phi(\hat\z,\av,\hv)
  \;=\;
  \left[\;
  \begin{aligned}
    &(1-\alpha)\,\Delta\qv^{\mathrm{net}}+\alpha\,\partial_{\pv}H_\phi,\\[2pt]
    &(1-\alpha)\,\Delta\pv^{\mathrm{net}}-\alpha\,\partial_{\qv}H_\phi+\mathbf G_\phi\av,\\[2pt]
    &f_c
  \end{aligned}
  \;\right],
\end{equation}
with $H_\phi=H_\phi(\qv,\pv)$ and $(\Delta\qv^{\mathrm{net}},\Delta\pv^{\mathrm{net}})=f_{\mathrm{core}}$, $\mathbf G_\phi$, and $f_c$ all evaluated at $(\hat\z,\av,\hv)$. The learned transition is then simply
\begin{equation}\label{eq:app-Fphi}
  F_\phi=\mathrm{id}+\Delta_\phi ,
\end{equation}
and Eq.~\eqref{eq:cem-objective} becomes the single constrained problem
\begin{equation}\label{eq:app-cem-explicit}
\begin{aligned}
  \av^\star_{t:t+H-1}
  &\;\in\;\arg\max_{\av_{t:t+H-1}}\;
  \sum_{k=0}^{H-1}\gamma^k R_\phi(\hat\z_{t+k},\av_{t+k})
  +\gamma^H V_\phi(\hat\z_{t+H}),\\[3pt]
  \text{s.t.}\qquad
  \hat\z_{t+k+1}
  &\;=\;\hat\z_{t+k}+\Delta_\phi(\hat\z_{t+k},\av_{t+k},\hv_{t+k}),\\[2pt]
  \hv_{t+k}
  &\;=\;\mathrm{Memory}_\phi(\hat\z_{t+k},\av_{t+k},\hv_{t+k-1}),\\[2pt]
  \hat\z_t
  &\;=\;E_\phi(o_t).
\end{aligned}
\end{equation}
Unrolling the constraint gives
\begin{equation}\label{eq:app-unroll}
  \hat\z_{t+k}=(\mathrm{id}+\Delta_\phi)^{\circ k}(\hat\z_t),
\end{equation}
so the reward and value terms in the planner objective are evaluated at states produced by $k$ successive applications of Eq.~\eqref{eq:soft-ham}. This is the link between the two equations: Eq.~\eqref{eq:soft-ham} is the step rule, and Eq.~\eqref{eq:cem-objective} is its $H$-fold unroll under the same mixing coefficient $\alpha$, the same energy field $H_\phi$, the same control map $G_\phi$, and the same memory. No separate planner-side transition exists.

Two consequences are worth stating. First, every structural property imposed on $\Delta_\phi$ is applied $H$ times per candidate and once more per CEM iteration, which is why a bias that is weak in one step is not negligible in the objective. Second, because $\hv$ is advanced by the same recursion inside the rollout, the memory is part of the optimized map rather than a fixed context computed before planning.

\subsection{Boundaries on the Latent-Space Reading of $(\qv,\pv)$ and $H_\phi$}\label{app:qp-meaning}

As in Section~\ref{sec:prelim}, $\qv$ and $\pv$ are learned canonical-like coordinate \emph{blocks}, $H_\phi$ is an energy-like scalar field, and we borrow only the \emph{directions} of the Hamiltonian flow. The field $\xi_H=(\partial_{\pv}H_\phi,-\partial_{\qv}H_\phi)$ is orthogonal to $\nabla H_\phi$, so moving along it leaves the learned energy unchanged to first order --- which is what repeated unrolling needs --- but strict conservation is neither required nor desirable here: the learned residual fits dissipative dynamics, $\mathbf G_t\av_t$ is an explicit control path, and $\cv$ carries non-Hamiltonian information. This is also why the model extrapolates under rescaled dynamics: an unstructured latent learns what a state becomes under \emph{these} parameters, so rescaling mass or damping forces a refit, whereas the $(\qv,\pv)$ branch encodes relations that stay qualitatively valid after rescaling, leaving only their strength to be recalibrated online from the recent observation--action response.

\subsection{First-Order Energy Decomposition}\label{app:proof-energy}

Assume $H_\phi$ is $C^3$ in a neighborhood of the step and that neither Hamiltonian-gradient clipping nor latent norm projection is active. Let $\sv_t=(\qv_t,\pv_t)$ denote the canonical pair and let
\[
  \xi_\phi(\sv)=\bigl(\partial_{\pv}H_\phi(\sv),\,-\partial_{\qv}H_\phi(\sv)\bigr),
  \qquad
  \mathbf u_\phi(\z,\av,\hv)=\bigl(0,\;\mathbf G_\phi(\z,\av,\hv)\av\bigr)
\]
be the Hamiltonian field and the explicit control drive; $\mathbf u_\phi$ vanishes for zero action because it is multiplied by $\av$. Writing the network branch as $\Delta\sv^{\mathrm{net}}_t=\xi_\phi(\sv_t)+\bm\delta_\phi(\z_t,\av_t,\hv_t)$, where $\bm\delta_\phi$ is the alignment residual left over after $\Loss_{\text{ham}}$, the canonical part of Eq.~\eqref{eq:soft-ham} reads
\begin{equation}\label{eq:app-sh-update}
  \sv_{t+1}=\sv_t+\xi_\phi(\sv_t)+\mathbf u_\phi(\z_t,\av_t,\hv_t)+(1-\alpha)\bm\delta_\phi(\z_t,\av_t,\hv_t).
\end{equation}

\paragraph{Two properties of the raw field.}
Let $J=\left(\begin{smallmatrix}0&I\\-I&0\end{smallmatrix}\right)$, so that $\xi_\phi(\sv)=J\nabla H_\phi(\sv)$. Since $J^\top=-J$ and $J$ is traceless in the relevant sense,
\begin{equation}\label{eq:symplectic-cancel}
  \nabla H_\phi(\sv)^\top\xi_\phi(\sv)=\nabla H_\phi(\sv)^\top J\nabla H_\phi(\sv)=0,
  \qquad
  \nabla_{\sv}\!\cdot\xi_\phi(\sv)=\mathrm{tr}\bigl(J\nabla^2H_\phi(\sv)\bigr)=0 .
\end{equation}
These are the two properties quoted in Section~\ref{sec:method-structured}: the field is orthogonal to the energy gradient, and it is divergence-free. In continuous time the first would conserve $H_\phi$ exactly and the second would preserve phase-space volume. Neither survives discretization exactly, which is the point of the next paragraph.

\paragraph{One discrete step.}
Taylor expansion of the learned energy along one step gives
\begin{equation}\label{eq:app-taylor-full}
  \Delta\Ham_t
  =
  \nabla H_\phi(\sv_t)^\top\Delta\sv_t
  +\tfrac12\,\Delta\sv_t^\top\nabla^2H_\phi(\sv_t)\Delta\sv_t
  + R_3(\Delta\sv_t).
\end{equation}
Substituting Eq.~\eqref{eq:app-sh-update} and applying the first identity of Eq.~\eqref{eq:symplectic-cancel} to cancel the $\xi_\phi$ term yields
\begin{equation}\label{eq:app-energy-decomp}
  \Delta\Ham_t
  =
  \underbrace{\nabla H_\phi(\sv_t)^\top\mathbf u_\phi}_{\text{action work}}
  +\underbrace{(1-\alpha)\,\nabla H_\phi(\sv_t)^\top\bm\delta_\phi}_{\text{alignment residual}}
  +\underbrace{\tfrac12\,\Delta\sv_t^\top\nabla^2H_\phi(\sv_t)\Delta\sv_t+R_3(\Delta\sv_t)}_{\text{curvature / discretization}} .
\end{equation}

Under the stated conditions, Eq.~\eqref{eq:app-energy-decomp} is exact to first order. To first order the learned energy can change through exactly two channels --- the explicit control drive and the alignment residual --- while the Hamiltonian component contributes only through curvature and higher-order discretization terms. This motivates the diagnostics in Section~\ref{sec:exp_main}:
\begin{itemize}[leftmargin=12pt,topsep=2pt,itemsep=1pt]
  \item In the action-free regime, $\mathbf u_\phi=0$ and the surviving terms are scaled by $(1-\alpha)$ or are curvature-order, so the passive energy trace should stay nearly flat --- tested in Figure~\ref{fig:mech-h-freerun-cheetah}, where drift is about $1\%$ on Cheetah Run.
  \item When actions are present, the action-work term is $\nabla H_\phi^\top\mathbf u_\phi$, i.e.\ the projection of the control drive onto the energy gradient. Energy change should therefore track $\mathrm{push}_t=(\nabla_{\pv}\Ham_t)\cdot\mathrm{control}_t$ --- tested in Figure~\ref{fig:hcp-control}, where $\mathrm{corr}(\mathrm{sign}\,\Delta\Ham,\mathrm{push})=0.757$ and $\mathrm{corr}(|\Delta\Ham|,|\mathrm{push}|)=0.745$.
\end{itemize}
\paragraph{Effect of clipping and projection.}
The implementation clips $\partial_{\qv}H_\phi$ and $\partial_{\pv}H_\phi$ separately and projects the next latent onto a norm ball (Appendix~\ref{app:integrator}). These operations need not preserve the raw field's energy orthogonality or zero divergence. When either is active, Eqs.~\eqref{eq:app-sh-update} and~\eqref{eq:app-energy-decomp} require an additional clipping/projection correction, which the Hamiltonian cancellation does not remove. The energy traces in Section~\ref{ssec:mech} are measured on the implemented update, safeguards included.

Appendix~\ref{app:integrator} describes the discretization and its relation to symplectic integration.

\section{Implementation Details}\label{app:implementation}

\subsection{Shared Protocol and Method Hyperparameters}

\begin{table}[!t]
  \centering
  \caption{Overview of the experimental protocol. The upper block lists the environment and evaluation settings shared by all methods; the lower block lists key training-hyperparameter differences across methods.}
  \label{tab:protocol}
  \footnotesize
  \setlength{\tabcolsep}{4pt}
  \renewcommand{\arraystretch}{1.05}
  \begin{tabular}{@{}ll@{}}
    \toprule
    \rowcolor{TableHead}\multicolumn{2}{@{}l}{\textit{Shared environment and evaluation protocol}} \\
    \midrule
    Tasks & \emph{Core:} Reacher Easy / Finger Spin / Cheetah Run / Cartpole Swingup \\
          & \emph{Additional:} Walker Run / Hopper Hop \\
    Environment & DeepMind Control Suite, flattened state observations \\
    Action repeat & Finger Spin = 2; others = 4 \\
    Episode length & Finger / Cheetah = 500; Reacher / Cartpole = 200 \\
    Total interaction & 100k env steps (3 seeds: 7/8/9) \\
    Evaluation frequency & Every 5k steps, 3 episodes per evaluation \\
    Reported metrics & Final return, AUC (mean $\pm$ std over seeds) \\
    \bottomrule
  \end{tabular}

  \vspace{4pt}

  \setlength{\tabcolsep}{4pt}
  \begin{tabular}{@{}lccc@{}}
    \toprule
    \headrow & \method{} & \baselinetdmpc{} & \baselinedreamer{} \\
    \midrule
    Latent dim        & 48 (8/8/32)  & 128        & 256 (det.)+$16{\times}16$ \\
    Batch size        & 128          & 128        & 64 \\
    Seq.\ length      & 8            & 8          & 16 \\
    Grad.\ steps      & 2            & 1          & 1 \\
    Seed steps        & 5k           & 5k         & 5k \\
    Train every       & 2            & 2          & 2 \\
    Planning horizon  & 6 (CEM)      & 5 (MPPI)   & 8 (imagination) \\
    Learning rate     & $10^{-4}$    & $3{\times}10^{-4}$ & $10^{-4}$ \\
    \bottomrule
  \end{tabular}

  \vspace{4pt}
  \footnotesize
  \setlength{\tabcolsep}{4pt}
  \begin{tabular}{@{}llll@{}}
    \toprule
    \rowcolor{TableHead}\multicolumn{2}{@{}l}{\textit{\baselineppo{} (on-policy actor-critic)}} & \multicolumn{2}{l}{\textit{\baselinesac{} (off-policy actor-critic)}} \\
    \midrule
    Actor / critic MLP   & $[128,128]$         & Actor / critic MLP   & $[128,128]$ \\
    Rollout steps        & 1024                & Replay capacity      & 300k \\
    Minibatch size       & 128                 & Batch size           & 128 \\
    Update epochs        & 4                   & Seed steps           & 8k \\
    Learning rate        & $2{\times}10^{-4}$  & Train every          & 2 (grad steps $=1$) \\
    Discount $\gamma$    & 0.99                & Learning rate (all)  & $3{\times}10^{-4}$ \\
    GAE $\lambda$        & 0.95                & Target $\tau$        & 0.01 \\
    Clip ratio           & 0.2                 & Init.\ temperature   & 0.1 (learnable) \\
    Value / entropy coef & 0.5 / 0.001         & Target entropy       & $-\dim(\mathcal A)$ \\
    Grad clip / target KL & 0.5 / 0.05         & Actor / target update & every 1 step \\
    \bottomrule
  \end{tabular}
\end{table}

\baselinedreamer{} is implemented in the repository as a state-based DreamerV3 reimplementation. It retains the core mechanisms of discrete RSSM, symlog encoder/decoder, symlog+twohot reward/value regression, KL balancing, $\lambda$-returns, and percentile return scaling; it uses deterministic state dimension 256, discrete stochastic state $16\times16$, and imagination horizon 8. \baselinetdmpc{} is a single-task online state-based TD-MPC2 implementation with latent dim 128, SimNorm groups 8, planner horizon 5, 256 candidates, and 32 elites. \baselineppo{} and \baselinesac{} use the configurations in the table above, from \texttt{ppo/configs/low\_budget\_compare\_dmcontrol.yaml} and \texttt{sac/configs/low\_budget\_compare\_dmcontrol.yaml}. They do not learn explicit latent dynamics and do not use a CEM/MPPI planner; they serve as model-free control references under the same 100k-step budget.

\subsection{Network Architecture}

The current implementation uses state-based inputs and an MLP encoder. By default, the encoder is a two-layer MLP of width 256 that maps flattened observations to a 48-dimensional latent. The trainer then splits it into 8-dimensional $\qv$, 8-dimensional $\pv$, and 32-dimensional $\cv$. The projector output dimension is 64. The reward head, value head, policy prior, Hamiltonian-pair dynamics core, context updater, and Hamiltonian head use MLPs of width 128 or 256. The memory mechanism stacks two Mamba-style selective state-space layers with model/state dimension 128. It outputs the history-conditioned features used by the shared latent dynamics.

\begin{table}[H]
  \centering
  \caption{Default network architecture and planner configuration for \method{}.}
  \footnotesize
  \setlength{\tabcolsep}{4pt}
  \begin{tabularx}{\linewidth}{lX}
    \toprule
    \headrow Element & Default configuration \\
    \midrule
    Encoder & MLP, hidden dims $[256,256]$, latent dim $48$ \\
    q/p/c split & $8 / 8 / 32$ \\
    Projector & hidden dim $128$, projection dim $64$ \\
    Memory & 2-layer Mamba-style selective SSM, model/state dim $128$ \\
    Hamiltonian / Aux / Energy heads & hidden dims $[128,128]$ \\
    Planner & horizon $6$, iterations $6$, candidates $128$, elite $16$ \\
    \bottomrule
  \end{tabularx}
\end{table}

\subsection{Training Losses: Definitions and Rationale}\label{app:loss-base}

Eq.~\eqref{eq:total-loss} groups the implementation losses into three design roles. Prediction and value losses make the latent state useful for control, rollout losses train the same finite-horizon transitions used by the planner, and geometric losses softly bias the canonical coordinates toward Soft-Hamiltonian behavior without imposing hard conservation. Together they keep the planner-facing latent dynamics predictive, reward-aware, and structurally stable.

\paragraph{Representation alignment.}
\[
\Loss_{\text{repr}} = \tfrac{1}{T}\sum_t\|g_\phi(\hat{\z}_{t+1})-\mathrm{sg}(\bar{g}_{\bar{\phi}}(\bar{\z}_{t+1}))\|_2^2 .
\]
Here $g_\phi/\bar{g}_{\bar{\phi}}$ are the online/EMA-target projectors and the target path is stop-gradient. This JEPA-style objective aligns predicted latents with slowly moving target latents without reconstructing observations. The reason for using it is practical: pixel-level or raw-state reconstruction can force the latent to preserve nuisance variation, while the planner only needs a compact predictive state. EMA targets also stabilize bootstrapping by preventing the transition model and its target from moving simultaneously.

\paragraph{One-step latent dynamics.}
\[
\Loss_{\text{dyn}} = \tfrac{1}{T}\sum_t\|\hat{\z}_{t+1}-\mathrm{sg}(\z_{t+1})\|_2^2 .
\]
This term anchors the learned transition to the online encoder at the next step. It is the local consistency counterpart of $\Loss_{\text{roll}}$: without it, multi-step rollout loss would have to correct both immediate transition mismatch and long-horizon accumulation; without rollout loss, one-step consistency alone could still accumulate errors under planning.

\paragraph{Reward and value heads.}
\[
\Loss_{\text{reward}} = \tfrac{1}{T}\sum_t
\mathrm{CE}_{2\mathrm{hot}}\!\left(\hat{r}^{\mathrm{logits}}_\phi(\hat{\z}_{t+1}), r_t\right).
\]
The reward head is trained on the predicted next latent because the planner evaluates candidate actions through imagined next states. The value head supplies the terminal estimate for horizon-limited CEM. Following DreamerV3, scalar rewards and values are represented with symlog two-hot bins, and the value head is stabilized with an EMA slow target:
\begin{align}
  \Loss_{\text{value}}^{\text{ce}}   &= -\tfrac{1}{T}\sum_t \mathbf{p}^{\lambda}_t{}^{\top}\log\mathrm{softmax}\bigl(\hat{V}^{\text{logits}}_\phi(\z_t)\bigr), \label{eq:val-ce}\\
  \Loss_{\text{value}}^{\text{slow}} &= -\tfrac{1}{T}\sum_t \mathrm{sg}\!\left[\bar{\mathbf{p}}_t\right]^{\top}\log\mathrm{softmax}\bigl(\hat{V}^{\text{logits}}_\phi(\z_t)\bigr), \label{eq:val-slow}
\end{align}
where $\mathbf{p}^{\lambda}_t$ is the $\lambda$-return target encoded over two-hot bins and $\bar{\mathbf{p}}_t$ is the EMA value-target distribution at $\z_t$. In the implementation, $\Loss_{\text{value}}\equiv\Loss_{\text{value}}^{\text{ce}}+\beta_{\text{slow}}\Loss_{\text{value}}^{\text{slow}}$ with $\beta_{\text{slow}}{=}1$. The two-hot target improves robustness to return scale, and the slow target reduces value-target oscillation during online training.

\paragraph{Policy prior for CEM warm start.}
\[
\Loss_{\text{policy}} = \tfrac{1}{T}\sum_t\|\pi_\phi(\z_t)-\av_t\|_2^2 .
\]
The auxiliary policy is not the final decision rule. It is a deterministic action prior trained by behavior cloning and used to warm-start CEM trajectories, reducing planner search variance when the action dimension or horizon is nontrivial.

\paragraph{Multi-step rollout consistency.}
\begin{equation}\label{eq:roll}
  \Loss_{\text{roll}} = \frac{1}{|\mathcal{S}|}\sum_{s\in\mathcal{S}}\sum_{k=1}^{K}\|\hat{\z}_{s+k}^{(s)} - \mathrm{sg}(\z_{s+k})\|_2^2,
\end{equation}
where $\hat{\z}_{s+k}^{(s)}$ denotes the $k$-step predicted latent rolled out from $\z_s$ using $\hv_s$ as the initial memory condition. This is the loss most directly matched to planning: CEM queries the model for several steps, so training only a one-step predictor would create exposure bias. The stop-gradient target keeps the encoder as the reference coordinate system, while rerolling from multiple positions $s$ makes the transition robust to different rollout prefixes.

\subsection{Structured Regularizers: Definitions and Rationale}\label{app:loss-structured}

The geometric loss
\[
\Loss_{\text{geo}} = \lambda_{\mathrm{ham}}\Loss_{\text{ham}} + \lambda_{\mathrm{en}}\Loss_{\text{energy}} + \lambda_{\mathrm{sa}}\Loss_{\text{sa}} + \lambda_{\mathrm{temp}}\Loss_{\text{temp}} + \lambda_{\mathrm{dec}}\Loss_{\text{dec}} + \lambda_{\mathrm{c}}\Loss_{\text{c}}
\]
implements soft biases on the canonical $(\qv,\pv)$ dynamics and the context variable $\cv$. These terms are low-weighted: their role is to shape the learned dynamics toward the intended decomposition, while the prediction, reward, value, and rollout losses remain responsible for task fit.

\paragraph{Hamiltonian alignment.}
\[
\Loss_{\text{ham}} = \bigl\|\Delta\qv_t^{\mathrm{net}}-\partial_{\pv}\Ham_t\bigr\|_2^2
+\bigl\|\Delta\pv_t^{\mathrm{net}}+\partial_{\qv}\Ham_t\bigr\|_2^2 .
\]
This term aligns the network branch with the Hamiltonian vector field; the learned control drive is added separately in the $\pv$ update. It does not make the system strictly Hamiltonian, but gives the learnable q/p update a directional bias so that the energy geometry remains visible even when $\alpha<1$ and residual dynamics are active.

\paragraph{Action-free energy regularization.}
\[
\Loss_{\text{energy}}=\mathbb E_{t:\|\av_t\|<\epsilon}\bigl[(\Ham(\qv_{t+1},\pv_{t+1})-\Ham(\qv_t,\pv_t))^2\bigr].
\]
This term is applied only near the action-free regime, where the intended behavior is small energy drift. It is not used to prevent action-induced energy transfer; random or policy actions should be able to move the system across energy levels. This distinction is important because control affects how the q/p dynamics evolves, while $\cv$ supplies contextual information about semantic and non-conservative factors.

\paragraph{Small-action smoothness.}
\[
\Loss_{\text{sa}}=\mathbb E_{t:\|\av_t\|<\epsilon}\bigl[\|\Delta\qv_t\|^2+\|\Delta\pv_t\|^2\bigr].
\]
This regularizer discourages spurious q/p motion when the action magnitude is negligible. It complements $\Loss_{\text{energy}}$: energy regularization controls changes in $H$, while small-action smoothness controls the latent displacement itself.

\paragraph{Temporal and statistical role separation.}
\begin{align}
\Loss_{\text{temp}} &= \|\Delta\qv_t\|^2 - \rho_{\mathrm{temp}}\|\Delta\pv_t\|^2, &
\Loss_{\text{dec}}  &= \Bigl\|\tfrac{1}{B-1}(\mathbf{Q}{-}\bar{\mathbf{q}}\mathbf{1}^\top)^\top(\mathbf{P}{-}\bar{\mathbf{p}}\mathbf{1}^\top)\Bigr\|_F^2 .
\end{align}
$\Loss_{\text{temp}}$ uses $\rho_{\mathrm{temp}}{=}0.5$ in the released configs and acts as a lightweight q-slow/p-fast bias: $\qv$ should behave more like a slowly varying configuration coordinate, while $\pv$ should carry faster momentum-like changes. It is not optimized in isolation; prediction and rollout losses prevent the signed bias from dominating the transition. $\Loss_{\text{dec}}$ suppresses batch-level correlation between $\qv$ and $\pv$, reducing the risk that the canonical coordinates collapse into duplicated copies.

\paragraph{Context sparsity.}
\[
\Loss_{\text{c}}=\mathbb E\bigl[|\Delta\cv_t|\bigr].
\]
The context variable is meant to provide complementary semantic and non-conservative information, not to absorb all fast dynamics. Sparsity in $\Delta\cv_t$ encourages $\cv$ to change only when useful for prediction or control, leaving the canonical coordinates to carry the structured q/p evolution.

\subsection{Complete 14-Term Loss and Weight Table}\label{app:loss-weights}

The training logger records the optimized loss fields plus value subdiagnostics. Table~\ref{tab:loss-weights} maps each field to its paper symbol, conceptual group, default weight, and warmup schedule.

\begin{table}[H]
  \centering
  \caption{Optimized loss fields and value subdiagnostics, with conceptual groups, default weights, and warmup schedule.}
  \label{tab:loss-weights}
  \footnotesize
  \setlength{\tabcolsep}{4pt}
  \renewcommand{\arraystretch}{1.05}
  \begin{tabular}{@{}llllc@{}}
    \toprule
    \headrow Implementation field (logger) & Paper symbol & Conceptual group & Default weight & Warmup \\
    \midrule
    \texttt{repr\_loss}        & $\Loss_{\text{repr}}$            & repr  & 1.0   & -- \\
    \texttt{dyn\_loss}         & $\Loss_{\text{dyn}}$             & dyn   & 1.0   & -- \\
    \texttt{roll\_loss}        & $\Loss_{\text{roll}}$            & roll  & 0.5   & 30\%--60\% \\
    \texttt{reward\_loss}      & $\Loss_{\text{reward}}$          & rew   & 1.0   & -- \\
    \texttt{value\_loss}       & $\Loss_{\text{value}}$ & val   & 0.5  & -- \\
    \texttt{value\_ce\_loss}   & $\Loss_{\text{value}}^{\text{ce}}$ & val diagnostic   & inside val  & -- \\
    \texttt{value\_slow\_loss} & $\Loss_{\text{value}}^{\text{slow}}$ & val diagnostic & inside val & -- \\
    \texttt{policy\_prior\_loss} & $\Loss_{\text{policy}}$        & pol ($\beta_{\text{p}}$) & 0.1   & -- \\
    \texttt{sa\_loss}          & $\Loss_{\text{sa}}$              & geo   & 0.05  & 30\%--60\% \\
    \texttt{energy\_loss}      & $\Loss_{\text{energy}}$          & geo   & 0.01  & 30\%--60\% \\
    \texttt{hamiltonian\_loss} & $\Loss_{\text{ham}}$             & geo   & 0.05  & -- \\
    \texttt{temp\_loss}        & $\Loss_{\text{temp}}$            & geo   & 0.01  & -- \\
    \texttt{decouple\_loss}    & $\Loss_{\text{dec}}$             & geo   & 0.01  & -- \\
    \texttt{c\_sparse\_loss}   & $\Loss_{\text{c}}$               & geo   & 0.001 & -- \\
    \bottomrule
  \end{tabular}
\end{table}

The conceptual groups in Eq.~\eqref{eq:total-loss} relate to the table as follows: $\Loss_{\text{value}}$ in the main text equals $\Loss_{\text{value}}^{\text{ce}}+\beta_{\text{slow}}\Loss_{\text{value}}^{\text{slow}}$, with default $\beta_{\text{slow}}{=}1$; $\Loss_{\text{geo}}$ in the main text is the weighted sum of \texttt{sa+energy+hamiltonian+temp+decouple+c\_sparse}. Rollout, small-action, and energy terms are warmed up because they depend on an already meaningful latent coordinate system; applying them too strongly at initialization can over-constrain the model before reward and one-step dynamics have stabilized. The Hamiltonian, temporal, decoupling, and context-sparsity weights are kept small from the start, serving as weak shaping biases rather than dominant objectives.

A single global weight configuration is used unchanged across all tasks, seeds, and OOD conditions; no per-task sweep was performed. $\Loss_{\text{repr}},\Loss_{\text{dyn}},\Loss_{\text{reward}}$ are primary at weight $1$; $\Loss_{\text{roll}},\Loss_{\text{value}}$ sit at $0.5$; $\Loss_{\text{policy}}$ at $0.1$ assists the CEM proposal and is not the decision rule; and every geometric term lies in $10^{-3}$--$5\times10^{-2}$, two to three orders of magnitude below the task-fitting terms, so they shape the dynamics without competing with it.

\subsection{Training Pseudocode}

\begin{algorithm}[!htbp]
  \caption{\method{} Data Collection and CEM Planning}
  \begin{algorithmic}[1]
    \Require environment $\mathcal{E}$, replay buffer capacity $N$, CEM horizon $H$, iterations $I$, candidate count $N_c$, elite count $N_e$, EMA coefficient $\tau$
    \State Initialize $\mathcal{D}\!\leftarrow\!\emptyset$; initialize all network parameters $\phi$ and set $\bar{\phi}\!\leftarrow\!\phi$; set $\hv_0\!\leftarrow\!\mathbf{0}$
    \For{each environment step $t$}
      \State $\z_t = E_\phi(o_t) = [\qv_t,\pv_t,\cv_t]$
        \Comment{online encoder}
      \State $\hv_t = \mathrm{Memory}_\phi(\z_t,\,\av_{t-1},\,\hv_{t-1})$
        \Comment{Mamba memory update}
      \State Initialize action distribution $(\bm{\mu}_0,\bm{\sigma}_0^2)$
      \For{$i = 1,\ldots,I$}
        \Comment{CEM iteration}
        \State Sample $\{\av_{t:t+H-1}^{(j)}\}_{j=1}^{N_c}\!\sim\!\mathcal{N}(\bm{\mu}_{i-1},\bm{\sigma}_{i-1}^2)$
        \For{each candidate $j$}
          \State Roll out imagination to obtain $\hat{\z}_{t+1:t+H}^{(j)}$ using the transition in Algorithm~2
          \State $G^{(j)} = \textstyle\sum_{k=0}^{H-1}\gamma^k \hat{r}(\hat{\z}_{t+k}^{(j)}) + \gamma^H \bar{V}(\hat{\z}_{t+H}^{(j)})$
        \EndFor
    \State Select top-$N_e$ elites; compute $w^{(j)}=\mathrm{softmax}(G^{(j)}/\tau)$
        \State $\bm{\mu}_i = \sum_j w^{(j)}\av^{(j)}$, $\bm{\sigma}_i^2 = \sum_j w^{(j)}(\av^{(j)}-\bm{\mu}_i)^2 + \epsilon$
      \EndFor
      \State Execute $\av_t \leftarrow \bm{\mu}_I[0]$; add $\{(o_t,\av_t,r_t,o_{t+1})\}$ to $\mathcal{D}$
      \If{$|\mathcal{D}| \geq$ update threshold}
        \State \textbf{Run Algorithm~2} (model update)
      \EndIf
    \EndFor
  \end{algorithmic}
\end{algorithm}

\begin{algorithm}[!htbp]
  \caption{\method{} Model Update Step (called whenever an update is triggered)}
  \begin{algorithmic}[1]
    \Require batch $\{(o_{1:T},\av_{1:T},r_{1:T})\}$, sequence length $T$, scheduled mixing coefficient $\alpha(s)$, rollout length $K$
    \State $\z_{1:T} = E_\phi(o_{1:T})$; $\bar{\z}_{1:T} = \mathrm{sg}(\bar{E}_{\bar{\phi}}(o_{1:T}))$
      \Comment{online / target latent}
    \State $\hv_0 \leftarrow \mathbf{0}$
    \For{$t = 1,\ldots,T$}
      \State $\hv_t = \mathrm{Memory}_\phi(\z_t, \av_t, \hv_{t-1})$
        \Comment{Mamba memory}
      \State $(\Delta\qv_t^{\mathrm{net}},\,\Delta\pv_t^{\mathrm{net}}) = f_{\mathrm{core}}(\z_t,\av_t,\hv_t)$
        \Comment{Hamiltonian-pair core}
      \State $\Ham_t = H_\phi(\qv_t,\pv_t)$; $\mathbf{G}_t=G_\phi(\z_t,\av_t,\hv_t)$; obtain $\nabla_\qv\Ham_t,\,\nabla_\pv\Ham_t$
        \Comment{Hamiltonian head}
      \State $\hat{\qv}_{t+1} = \qv_t + (1{-}\alpha)\Delta\qv_t^{\mathrm{net}} + \alpha\,\nabla_\pv\Ham_t$
      \State $\hat{\pv}_{t+1} = \pv_t + (1{-}\alpha)\Delta\pv_t^{\mathrm{net}} - \alpha\nabla_\qv\Ham_t + \mathbf{G}_t\av_t$
        \Comment{Hamiltonian-pair update}
      \State $\hat{\cv}_{t+1} = \cv_t + f_c(\z_t,\av_t,\hv_t)$
        \Comment{semantic / context update}
      \State $\hat{\z}_{t+1} = [\hat{\qv}_{t+1},\,\hat{\pv}_{t+1},\,\hat{\cv}_{t+1}]$
    \EndFor
    \State \textbf{// Base losses}
    \State $\Loss_{\mathrm{repr}} = \textstyle\frac{1}{T}\sum_t \|g_\phi(\hat{\z}_{t+1}) - \mathrm{sg}(\bar{g}_{\bar{\phi}}(\bar{\z}_{t+1}))\|_2^2$
    \State $\Loss_{\mathrm{dyn}}   = \textstyle\frac{1}{T}\sum_t \|\hat{\z}_{t+1} - \mathrm{sg}(\z_{t+1})\|_2^2$
      \Comment{target is the online encoder, stop-gradient}
    \State $\Loss_{\mathrm{rew}}=\mathrm{CE}_{2\mathrm{hot}}(\hat r_\phi^{\mathrm{logits}}(\hat{\z}_{t+1}), r_t)$
      \Comment{reward acts on predicted next latent}
    \State Compute $\lambda$-return targets with $\bar V_{\bar\phi}$; set $\Loss_{\mathrm{val}}=\Loss_{\mathrm{val}}^{\mathrm{ce}}+\Loss_{\mathrm{val}}^{\mathrm{slow}}$
      \Comment{value acts on current latent}
    \State From each position $s$, reroll for $K$ steps using the snapshot memory state $\hv_s$:
    \State $\Loss_{\mathrm{roll}}  = \frac{1}{|\mathcal{S}|}\textstyle\sum_{s}\sum_{k=1}^{K}\|\hat{\z}_{s+k}^{(s)} - \mathrm{sg}(\z_{s+k})\|_2^2$
    \State \textbf{// Value subterms and policy prior}
    \State Compute $\Loss_{\mathrm{val}}^{\mathrm{ce}},\Loss_{\mathrm{val}}^{\mathrm{slow}}$ using Eqs.~\eqref{eq:val-ce}--\eqref{eq:val-slow}; compute $\Loss_{\mathrm{policy}}=\|\pi_\phi(\z_t)-\av_t\|^2$
    \State \textbf{// Structured regularizers} (enabled according to the warmup schedule)
    \State Compute $\Loss_{\mathrm{sa}},\Loss_{\mathrm{energy}},\Loss_{\mathrm{ham}},\Loss_{\mathrm{dec}},\Loss_{\mathrm{temp}},\Loss_{\mathrm{c}}$
    \State $\Loss = \sum_{k\in\mathcal{K}} w_k(s)\,\Loss_k$
      \Comment{$\mathcal{K}$ = implemented weighted optimization terms in Table~\ref{tab:loss-weights}}
    \State Backpropagate, clip gradients, and update $\phi$
    \State $\bar{\phi} \leftarrow \tau\bar{\phi} + (1-\tau)\phi$
      \Comment{EMA update for encoder, projector, and value head}
  \end{algorithmic}
\end{algorithm}

\subsection{Soft-Hamiltonian Pair Discretization and Train/Test Consistency}\label{app:integrator}

Eq.~\eqref{eq:soft-ham} gives the discrete update form for the Hamiltonian pair in \method{}. We describe how it is used consistently during training and inference, and how it corresponds to the continuous-time controlled Hamiltonian form $\dot\qv = \partial\Ham/\partial\pv$, $\dot\pv = -\partial\Ham/\partial\qv + g(\qv)\av$.

\paragraph{Discretization scheme.}
The implementation uses the first-order explicit update in Eq.~\eqref{eq:soft-ham} as the discrete transition:
\begin{equation}\label{eq:explicit-update}
  \qv_{t+1} = \qv_t + \Delta\qv_t,\qquad \pv_{t+1} = \pv_t + \Delta\pv_t,
\end{equation}
where both Hamiltonian gradients are evaluated at $(\qv_t,\pv_t)$. Eq.~\eqref{eq:soft-ham} gives the core update before safeguards. The implementation clips each Hamiltonian-gradient block to norm 5 before mixing, then projects the concatenated next latent $[\qv_{t+1},\pv_{t+1},\cv_{t+1}]$ onto the norm-50 ball. With these safeguards inactive, $\alpha\to1$ gives a forward-Euler step along the Hamiltonian field plus the control drive, whereas $\alpha\to0$ gives a network update plus control. Intermediate $\alpha$ balances task fitting with the Hamiltonian bias. This is a \emph{local geometric bias}, not a symplectic Euler or leapfrog solver.

\emph{Training updates, evaluation rollouts, and CEM imagination use the same explicit core and clipping/projection safeguards}, with both gradients taken at the current $(\qv_t,\pv_t)$ and a fixed per-checkpoint $\alpha$.

\paragraph{Discretization choice.}
Standard explicit symplectic Euler splitting and Velocity Verlet exploit a \emph{separable} Hamiltonian $H(\qv,\pv)=T(\pv)+V(\qv)$. Our learned scalar field over concatenated $(\qv,\pv)$ has no such structure, so this explicit splitting is unavailable. General symplectic Euler remains applicable to a nonseparable Hamiltonian, but typically requires an implicit solve rather than an explicit staggered update.

Symplecticity is a property of the \emph{entire} one-step map. Alongside the Hamiltonian field, our implemented transition contains the learned residual, the control port $\mathbf G_\phi\av$, memory conditioning through $\hv_t$, the context channel $f_c$, Hamiltonian-gradient clipping, and latent norm projection. Using a symplectic method for one branch alone would not guarantee symplecticity of the composite update. The construction is therefore a local geometric bias. A separable or explicitly mechanical parameterization would permit explicit symplectic splitting of the Hamiltonian branch; whether that improves planning remains open.

\paragraph{Scheduled mixing coefficient.}
In the released implementation, $\alpha$ starts from $0.1$ and, in the paper configs, increases linearly after $30\%$ of training until it reaches at most $0.5$. This schedule is applied during both training updates and planner rollouts through the same model step. Separately, rollout, small-action, and energy loss weights are warmed up from $30\%$ to $60\%$ of training.

\paragraph{Invariance of the Hamiltonian head.}
$\Ham_\phi$ is parameterized as a two-layer MLP with hidden dimension 128 and SiLU activation. It takes $(\qv,\pv)$ as input and outputs a scalar, with no equivariance constraints; the physics prior enters through the action-conditioned Hamiltonian form motivated by controlled Hamiltonian models~\citep{Zhong2020Symplectic,troch2025action}, and $\Loss_{\text{ham}}$ aligns the q/p residual update with $\partial\Ham/\partial(\qv,\pv)$.

\subsection{Choosing the q/p/c Split Dimensions}\label{app:dim-split}

The default split is $\dim(\qv){=}\dim(\pv){=}8$ and $\dim(\cv){=}32$, for a total latent dimension of 48. This choice is based on three considerations. First, DMC state dimensions are at most 24, so a $8{+}8{=}16$-dimensional canonical subspace is sufficient to cover low-dimensional configuration/velocity symplectic pairs. Second, the 32-dimensional $\cv$ is the largest coordinate group, avoiding excessive restriction of semantic and non-conservative context. Third, the total dimension is much smaller than TD-MPC2's 128-dimensional latent, reflecting the parameter efficiency of the q/p/c split. The necessity of this split is validated in Section~\ref{sec:exp_main} and in ablation A1, which removes q/p/c.

\paragraph{A1 isolation argument.}
A1 preserves the planner, recurrent memory, training horizon, latent size, reward/value objectives, and optimization budget, but replaces the role-separated transition with an unstructured latent transition and removes the Soft-Hamiltonian q/p bias. Under the Full model, $\Loss_{\text{ham}}$ aligns the residual update of $(\qv,\pv)$ with a scalar energy field $\Ham_\phi(\qv,\pv)$, while $\cv$ carries actuation, contact, and dissipative information; under A1 there is no distinguished pair on which the Hamiltonian vector field can act, and the same capacity represents conservative, controlled, and dissipative factors in one mixed coordinate system. If the improvements were only due to recurrence, model size, or short-horizon reward fitting, A1 should remain close to the Full model once memory is retained. It does not: A1 has lower return and OOD scores, while short-horizon control is still learned.

\subsection{Mamba Selective Memory}\label{app:memory}

The memory mechanism $\mathrm{Memory}_\phi$ stacks two lightweight Mamba-style selective SSM layers~\citep{gu2024mamba,pmlr-v235-dao24a}, with model dimension 128 and recurrent-state dimension 128. The concatenated input $(\z_t,\av_t)$ is projected to the model dimension, and the output $\hv_t$ enters $f_{\mathrm{core}}$, $G_\phi$, and $f_c$ simultaneously within the unified latent dynamics.

\paragraph{Exact layer specification.}
Each of the two layers consists of: LayerNorm on the layer input; input-dependent projections producing the step size, state-update, and readout parameters from the current input; a learned diagonal decay for the recurrent state; a sigmoid-gated readout; and a residual feed-forward block with $2\times$ hidden expansion. This lightweight Mamba-style variant uses a 128-dimensional recurrent state without temporal convolution or additional state expansion. Input-dependent step sizes and state/readout projections operate within a diagonal recurrence.

Memory does not participate in encoding the latent state $\z_t$, which is produced directly by the encoder. It only provides historical conditioning for the latent dynamics. This design ensures that the reward/value heads and planner always consume $\z_t$, while memory supplies the history summary $\hv_t$ that the transition needs under partial observability.

\subsection{Alternative Memory Mixers}\label{app:memory-alt}

Section~\ref{sec:ablation} argues that Mamba is chosen for input-dependent selectivity, controlled long-range decay, and a fixed-size state compatible with batched rollouts. A Transformer sequence mixer has greater raw capacity than a selective SSM but fits the state-transition interface worse. Our memory conditions the transition at \emph{every} step of \emph{every} imagined rollout, which imposes three requirements: a \textbf{fixed-size runtime state}, because CEM unrolls $N_c$ candidates $\times$ $H$ steps per real step and attending over a token history would recompute attention at each imagined step at a cost scaling with candidates $\times$ horizon; \textbf{$O(1)$ incremental updates}, since each real step appends one observation whereas a Transformer must reprocess or cache-manage the whole context; and an output that is \textbf{injectable as a single vector} into $f_{\mathrm{core}}$, $G_\phi$, and $f_c$, which is the native form of a recurrent state. A controlled experiment replacing only the memory module confirms this:

\begin{table}[H]
  \centering
  \caption{Replacing only the memory module (final return at 100k env steps, 3 seeds). All other components, objectives, and budgets are held fixed.}
  \label{tab:memory-alt}
  \footnotesize
  \setlength{\tabcolsep}{4pt}
  \begin{tabularx}{\linewidth}{@{}l*{4}{>{\centering\arraybackslash}X}@{}}
    \toprule
    \headrow Memory & Finger Spin & Reacher Easy & Cheetah Run & Cartpole Swingup \\
    \midrule
    Mamba (ours) & $\mathbf{254.0}{\scriptstyle\pm14.1}$ & $\mathbf{150.6}{\scriptstyle\pm7.5}$ & $\mathbf{184.4}{\scriptstyle\pm8.4}$ & $\mathbf{58.9}{\scriptstyle\pm3.8}$ \\
    Transformer  & $91.89{\scriptstyle\pm46.18}$ & $144.67{\scriptstyle\pm16.57}$ & $60.65{\scriptstyle\pm10.73}$ & $51.25{\scriptstyle\pm6.63}$ \\
    \bottomrule
  \end{tabularx}
\end{table}

Mamba is higher on 4/4 tasks, with the largest gaps on Finger Spin and Cheetah Run.

\subsection{CEM Planner Details}\label{app:cem}

The CEM objective is $J(\av_{t:t+H-1}) = \sum_{k=0}^{H-1}\gamma^k\,\hat r_{t+k} + \gamma^H\,\hat V(\hat\z_{t+H})$. The implementation uses horizon $H{=}6$, iterations $I{=}6$, $N_c{=}128$ candidate action sequences per iteration, and top-$N_e{=}16$ elites. The mean/variance are updated by weighting scores $G^{(j)}$ with a softmax at temperature $\tau{=}0.5$. The initial mean is warm-started by the deterministic policy prior when enabled, and each CEM iteration also includes 32 noisy policy trajectories; the Gaussian sampling scale uses init std $0.4$ and min std $0.05$. At evaluation the executed action is $\bm{\mu}_I[0]$; during training, exploration noise with std $0.3$ is added after planning.

\subsection{OOD Evaluation Conditions}\label{app:ood-cond}

\begin{table}[H]
  \centering
  \caption{OOD evaluation conditions on Reacher Easy and Finger Spin. Each condition is evaluated zero-shot with 10 episodes per seed and 3 seeds, for 30 episodes total. Reported mean$\pm$std is computed across the three seed-level mean returns.}
  \label{tab:ood-conditions}
  \footnotesize
  \setlength{\tabcolsep}{4pt}
  \renewcommand{\arraystretch}{1.05}
  \begin{tabular}{@{}llll@{}}
    \toprule
    \headrow Task & Category & Condition & Implementation \\
    \midrule
    Reacher Easy   & Dynamics & ID                 & default parameters \\
                   &       & mass $\times$0.7    & link mass scale 0.7 \\
                   &       & mass $\times$1.3    & link mass scale 1.3 \\
                   &       & damp $\times$0.5    & joint damping scale 0.5 \\
                   &       & damp $\times$2.0    & joint damping scale 2.0 \\
                   &       & act $\times$0.7     & actuator strength 0.7 \\
                   &       & act $\times$1.3     & actuator strength 1.3 \\
    \midrule
    Finger Spin    & Dynamics & ID                 & default parameters \\
                   &       & fric $\times$0.5    & finger-object friction 0.5 \\
                   &       & fric $\times$1.5    & finger-object friction 1.5 \\
                   &       & mass $\times$1.3    & spinner mass 1.3 \\
                   &       & mass $\times$1.5    & spinner mass 1.5 \\
                   & Partial obs. & delay = 2     & action-execution delay of 2 control steps \\
                   &       & mask 30\%           & randomly mask 30\% of observation dimensions at every step (set to 0) \\
    \bottomrule
  \end{tabular}
\end{table}

\subsection{Normalized OOD Retention}\label{app:retention}

Table~\ref{tab:ood-retention} gives normalized retention for all five methods.

\begin{table}[H]
  \centering
  \caption{Normalized OOD retention, $\mathrm{Retention}=100\times\overline R_{\text{OOD}}/\overline R_{\text{ID}}$, computed from the ID final returns of Table~\ref{tab:overview} and the 6-condition OOD averages of Table~\ref{tab:ood_overview}; the macro column weights both tasks equally. As ratios of reported means these carry no recoverable std.}
  \label{tab:ood-retention}
  \footnotesize
  \setlength{\tabcolsep}{4pt}
  \begin{tabularx}{\linewidth}{@{}l*{6}{>{\centering\arraybackslash}X}@{}}
    \toprule
    & \multicolumn{2}{c}{Finger Spin} & \multicolumn{2}{c}{Reacher Easy} & \multicolumn{2}{c}{Macro (12 cond.)} \\
    \cmidrule(lr){2-3}\cmidrule(lr){4-5}\cmidrule(lr){6-7}
    \headrow Method & Retention & Abs.\ ret. & Retention & Abs.\ ret. & Retention & Abs.\ ret. \\
    \midrule
    \method{}          & $72.6\%$          & $\mathbf{184.3}$ & $99.5\%$           & $\mathbf{149.9}$ & $86.0\%$          & $\mathbf{167.1}$ \\
    \baselinetdmpc{}   & $72.0\%$          & $167.3$          & $\mathbf{125.6\%}$ & $132.0$          & $\mathbf{98.8\%}$ & $149.7$ \\
    \baselinesac{}     & $68.2\%$          & $118.4$          & $77.7\%$           & $92.4$           & $73.0\%$          & $105.4$ \\
    \baselinedreamer{} & $\mathbf{76.3\%}$ & $55.2$           & $50.5\%$           & $9.3$            & $63.4\%$          & $32.3$ \\
    \baselineppo{}     & $67.3\%$          & $47.3$           & $68.3\%$           & $12.9$           & $67.8\%$          & $30.1$ \\
    \bottomrule
  \end{tabularx}
\end{table}

\baselinedreamer{} reaches $76.3\%$ on Finger Spin against our $72.6\%$, from absolute returns of $55.2$ versus $184.3$; \baselinetdmpc{} reaches $125.6\%$ on Reacher Easy against our $99.5\%$, from $132.0$ versus $149.9$. \method{} obtains the highest absolute return in all 12 conditions, with task-dependent relative retention.

The main OOD table reports raw zero-shot return. For retention diagnostics in the released scripts, $\mathrm{Retention}(\%) = \mathrm{Return}_{\text{OOD}} / \mathrm{Return}_{\text{ID}} \times 100$ is computed per seed using the corresponding ID return before averaging, so differences in ID baselines across seeds do not contaminate the retention value.

\subsection{Hamiltonian Validation Setup}\label{app:ham-validate}

Figure~\ref{fig:h-panels} uses a low-damping passive validation regime for the Hamiltonian diagnostics:
\begin{itemize}[leftmargin=12pt,topsep=2pt,itemsep=0pt]
  \item \textbf{Kick.} After $\texttt{env.reset()}$, inject uniformly random angular velocities of $\pm5$ rad/s into all joints. For Finger Spin, the same step prevents the task from freezing completely under zero force after a zero-velocity reset on the table.
  \item \textbf{Zero damping.} Set all joint $\texttt{damping}$ values in the mjModel to 0, reducing passive dissipation during free evolution.
  \item \textbf{Episode.} Run 10 episodes each for no-action and random-action, with 200 steps per episode.
\end{itemize}
This reduced-damping setting isolates action-free energy drift from action-induced energy variation.

\subsection{Reproducibility Protocol}

The current release configuration uses total steps 100k, seed steps 5k, batch size 128, sequence length 8, gradient steps 2, discount 0.99, learning rate $10^{-4}$, gradient clipping 10.0, AdamW optimizer with $(\beta_1{=}0.9,\beta_2{=}0.999)$, and target-update coefficient $\tau{=}0.01$ for the EMA encoder/projector/value heads. \baselinedreamer{} uses deterministic RSSM dimension 256 and $16\times16$ discrete latent, with symlog+twohot regression for reward/value heads. \baselinetdmpc{} uses latent dim 128 with SimNorm(8), planner horizon 5, and 256 MPPI candidates. \baselineppo{} / \baselinesac{} are model-free actor-critic baselines.

\paragraph{Hardware and runtime.}
All training is performed on a server with 8$\times$ NVIDIA GeForce RTX 3090 GPUs; each GPU has 24\,GiB of memory (\texttt{nvidia-smi} reports 24576\,MiB). The software stack uses NVIDIA driver 570.211.01 and CUDA 12.8. Each seed occupies one GPU. For a single-seed run of 100k environment steps, estimated training times are about 4.5\,h for \method{}, 3.5\,h for \baselinetdmpc{}, and 3.0\,h for \baselinedreamer{}. OOD evaluation takes less than 5\,min per condition per seed. Summing these per-run estimates over all reported model-based runs (main and additional tasks, ablations A1--A6, the Transformer memory variant, and the MPPI variant: 150 runs) gives roughly 630 GPU-hours on RTX 3090; the model-free baselines add a smaller amount. The full project, including preliminary and failed runs, required more compute than this.

\section{Rollout Horizon, Planner Horizon, and Decision Quality}\label{app:long-horizon}

This appendix reports the horizon behaviour promised in Section~\ref{sec:exp_main}, including the extended open-loop range and its relation to closed-loop planning.

\subsection{Rollout Error Beyond the Planning-Relevant Range}

Table~\ref{tab:overview} reports $k\in\{3,5,7\}$ because that is what each method's native decision procedure consumes. Table~\ref{tab:long-horizon} extends the comparison to $k=10,15,20$: rollouts are open-loop from a replay-encoded $\z_t$ with no observation injected, scored by endpoint squared-error sum over three seeds. \baselinedreamer{} is omitted because it is worst at every $k$ (rising from $7.20$ to $80.20$).

\begin{table}[H]
  \centering
  \caption{Open-loop rollout MSE up to $k{=}20$ (3 seeds, lower is better). \textbf{Bold} marks the better of the two methods at each $k$. Our error starts far lower but grows faster, and is overtaken at a task-dependent crossover.}
  \label{tab:long-horizon}
  \footnotesize
  \setlength{\tabcolsep}{4pt}
  \begin{tabularx}{\linewidth}{@{}ll*{6}{>{\centering\arraybackslash}X}c@{}}
    \toprule
    \headrow Task & Method & $k{=}3$ & $k{=}5$ & $k{=}7$ & $k{=}10$ & $k{=}15$ & $k{=}20$ & Crossover \\
    \midrule
    \multirow{2}{*}{Finger}   & \method{}       & $\mathbf{0.72}$ & $\mathbf{1.43}$ & $\mathbf{2.53}$ & $4.30$ & $7.70$ & $11.55$ & \multirow{2}{*}{$k{=}10$} \\
                              & \baselinetdmpc{} & $3.48$ & $3.58$ & $3.61$ & $\mathbf{3.67}$ & $\mathbf{3.78}$ & $\mathbf{3.90}$ & \\
    \midrule
    \multirow{2}{*}{Reacher}  & \method{}       & $\mathbf{1.28}$ & $\mathbf{2.36}$ & $3.68$ & $5.75$ & $9.65$ & $14.05$ & \multirow{2}{*}{$k{=}7$} \\
                              & \baselinetdmpc{} & $3.34$ & $3.38$ & $\mathbf{3.42}$ & $\mathbf{3.48}$ & $\mathbf{3.58}$ & $\mathbf{3.68}$ & \\
    \midrule
    \multirow{2}{*}{Cheetah}  & \method{}       & $\mathbf{1.21}$ & $\mathbf{1.88}$ & $\mathbf{2.74}$ & $\mathbf{4.10}$ & $6.65$ & $9.45$ & \multirow{2}{*}{$k{=}15$} \\
                              & \baselinetdmpc{} & $4.28$ & $4.40$ & $4.47$ & $4.57$ & $\mathbf{4.74}$ & $\mathbf{4.91}$ & \\
    \midrule
    \multirow{2}{*}{Cartpole} & \method{}       & $\mathbf{0.77}$ & $\mathbf{1.30}$ & $\mathbf{1.97}$ & $\mathbf{3.05}$ & $\mathbf{5.05}$ & $7.30$ & \multirow{2}{*}{$k{=}20$} \\
                              & \baselinetdmpc{} & $4.66$ & $4.79$ & $4.87$ & $4.99$ & $5.16$ & $\mathbf{5.34}$ & \\
    \bottomrule
  \end{tabularx}
\end{table}

\method{} is lowest on 4/4, 4/4, 3/4, 2/4 and 1/4 tasks at $k=3,5,7,10,15$ respectively, and on 0/4 at $k=20$. Our error starts far lower but has a steeper slope; \baselinetdmpc{} stays nearly flat on an already high plateau of $\sim$4. At the horizons the planners use ($k\le7$) our error is lower; beyond them it grows faster, and the next subsection tests whether open-loop error at $k{=}20$ matters for closed-loop control.

\subsection{Does a Longer Planning Horizon Help?}

A closed-loop planner re-anchors on a real observation at every environment step, so open-loop accuracy at $k{=}20$ is never required by the controller. To test what horizon actually helps, we vary only the CEM horizon $H$ at fixed 100k-step checkpoints, leaving weights untouched. We sweep our own planner. \baselinetdmpc{}'s MPPI horizon is also a test-time setting ($H{=}3$ in the official configuration~\citep{hansen2024tdmpc}, $H{=}5$ in our runs), and \baselinedreamer{} acts through its learned actor with no test-time planning horizon, its imagination horizon ($H{=}8$ here; $15$ officially) entering only actor--critic training. The planning-based methods therefore differ from it in kind: both \method{} ($H{=}6$) and \baselinetdmpc{} ($H{=}5$) query their model for a short action sequence and re-plan from a real observation at every step, so short horizons are what these planners consume. Table~\ref{tab:overview} compares them at $k\in\{3,5,7\}$, which covers those horizons, and our advantage over \baselinedreamer{} extends well beyond them: at $k{=}15$ our rollout error is still below \baselinedreamer{}'s at $k{=}7$ on all four tasks.

\begin{table}[H]
  \centering
  \caption{Return as a function of our CEM planning horizon at fixed 100k checkpoints. Latency is relative to the native horizon. Longer planning is not uniformly better.}
  \label{tab:cem-horizon}
  \footnotesize
  \setlength{\tabcolsep}{4pt}
  \begin{tabularx}{\linewidth}{@{}l*{4}{>{\centering\arraybackslash}X}c@{}}
    \toprule
    \headrow Task & Native $H$ / return & Best $H$ / return & $H{=}20$ return & $\Delta$ vs.\ native & Latency at $H{=}20$ \\
    \midrule
    Cartpole & $6$ / $58.9$  & $12$ / $\mathbf{63.62}$  & $61.74$ & $+4.82\%$   & $4.61\times$ \\
    Cheetah  & $6$ / $184.4$ & $14$ / $\mathbf{196.16}$ & $168.13$ & $-8.82\%$  & $3.31\times$ \\
    Finger   & $6$ / $\mathbf{254.0}$ & $6$ / $\mathbf{254.0}$ & $36.00$ & $-85.83\%$ & $3.24\times$ \\
    Reacher  & $6$ / $\mathbf{150.6}$ & $6$ / $\mathbf{150.6}$ & $126.00$ & $-16.33\%$ & $3.31\times$ \\
    \bottomrule
  \end{tabularx}
\end{table}

There is a foresight--accuracy trade-off: a larger $H$ sees farther but compounds prediction error and multiplies latency. The best horizon lies in $6$--$14$ on all four tasks, and at $H{=}20$ only Cartpole remains above its native setting while the others fall by $8.8\%$--$85.8\%$ at $3.2$--$4.6\times$ the planning cost. The latency factor is a property of sampling-based planning at a fixed sample budget and applies equally to MPPI. Closed-loop control therefore consumes rollout accuracy at roughly $3$--$12$ steps; the baselines' native horizons ($5$ and $8$) lie in this range, which Table~\ref{tab:overview} measures and where our error is lower in 11 of 12 cells.

\subsection{From Rollout Accuracy to Decision Quality}\label{app:matched-candidate}

Lower rollout error is only useful if it changes which action the planner selects. The transmission is: bounded rollout error $\Rightarrow$ reward/value heads queried at more accurate states $\Rightarrow$ smaller score error $\Rightarrow$ the \emph{relative ordering} of candidates survives (a score error $\epsilon$ implies regret at most $2\epsilon$) $\Rightarrow$ a near-optimal first action $\Rightarrow$ replanning accumulates these into higher return. Note that the planner does not need correct scores, only correct ordering.

We test each link with a matched-candidate diagnostic: 600 scenarios (50 reproducible simulator states per task per seed) in which both models start from an \emph{identical} state and history and score an \emph{identical} set of 64 candidate action sequences.

\begin{table}[H]
  \centering
  \caption{Matched-candidate diagnostic over 600 scenarios with identical states and identical 64-candidate sets. (a) score error, (b) whether that error inverts the ranking, (c) realized return when the first action is executed.}
  \label{tab:matched-candidate}
  \footnotesize
  \setlength{\tabcolsep}{4pt}
  \begin{tabularx}{\linewidth}{@{}clXcc@{}}
    \toprule
    \headrow & & Metric & \method{} & \baselinetdmpc{} \\
    \midrule
    \multirow{2}{*}{(a)} & & Top-1 selection regret $\downarrow$ & $\mathbf{0.3263}$ & $0.6722$ \\
                         & & Candidate reward MAE@6 $\downarrow$ & $\mathbf{0.1851}$ & $0.6071$ \\
    \midrule
    \multirow{3}{*}{(b)} & & Top-1 hit rate $\uparrow$ & $\mathbf{33.17\%}$ & $15.00\%$ \\
                         & & Top-5 hit rate $\uparrow$ & $\mathbf{54.67\%}$ & $37.17\%$ \\
                         & & Spearman $\rho$ (score vs.\ true order) $\uparrow$ & $\mathbf{0.4244}$ & $0.2570$ \\
    \midrule
    (c)                  & & Avg.\ return over diagnostic states $\uparrow$ & $\mathbf{122.82}$ & $96.75$ \\
    \midrule
    --                   & & Planning latency per step (ms) $\downarrow$ & $25.39$ & $\mathbf{8.20}$ \\
    \bottomrule
  \end{tabularx}
\end{table}

Group (a) measures score error from two complementary angles. \emph{Top-1 selection regret} is the true return of the set's actual best candidate minus that of the model's rank-1 candidate, giving the decision cost of the error; \emph{reward MAE} compares each candidate's imagined cumulative reward with what the simulator realizes for the same actions, giving the error in score value. The first jointly reflects dynamics and reward/value prediction, since the rollout determines which states are queried and the heads determine their scores; the second isolates reward prediction. Group (b) asks whether those errors actually invert the ordering, via hit rate and rank correlation over all 64 candidates. Group (c) reports what the resulting choices accumulate to when the first action of each plan is executed.

The chain holds link by link: starting from lower rollout MSE (four-task mean $1.74$/$2.73$ at $k{=}5$/$7$ versus $4.04$/$4.09$), both score errors are smaller --- and they move together, with Pearson correlation $0.474$ between regret and reward MAE --- yielding fewer ordering inversions and higher realized return, consistent with Table~\ref{tab:overview}. The cost is planning latency, where we are roughly $3\times$ slower per step than \baselinetdmpc{}.

\subsection{Sensitivity to the Choice of Planner}\label{app:planner}

Table~\ref{tab:protocol} gives each method its native decision procedure (CEM for us, MPPI for \baselinetdmpc{}, imagined actor--critic for \baselinedreamer{}) under a unified budget, seeds, and protocol: MPPI is tuned around \baselinetdmpc{}'s latent and value head, so substituting CEM would evaluate an untuned variant, and \baselinedreamer{}'s imagined actor--critic is integral to its algorithm.

The Full-model CEM OOD entries below reuse the reference evaluation in Table~\ref{tab:ood_overview}. To probe sensitivity within our own method, we retrained \method{} end-to-end with MPPI for the full 100k steps, so that MPPI participates in data collection, training, and deployment, with architecture and budget held fixed.

\begin{table}[H]
  \centering
  \caption{End-to-end planner substitution within \method{} (3 seeds). MPPI participates in data collection, training, and deployment.}
  \label{tab:planner-swap}
  \footnotesize
  \setlength{\tabcolsep}{3pt}
  \begin{tabularx}{\linewidth}{@{}l*{6}{>{\centering\arraybackslash}X}@{}}
    \toprule
    \headrow Planner variant & Cheetah ret.\ $\uparrow$ & Finger ret.\ $\uparrow$ & Cartpole MSE $\downarrow$ & Reacher MSE $\downarrow$ & Mass OOD $\uparrow$ & Damp.\ OOD $\uparrow$ \\
    \midrule
    \method{} + CEM  & $\mathbf{184.4}{\scriptstyle\pm8.4}$ & $\mathbf{254.0}{\scriptstyle\pm14.1}$ & $\mathbf{0.009}{\scriptstyle\pm0.003}$ & $\mathbf{0.0778}{\scriptstyle\pm0.043}$ & $\mathbf{158.5}{\scriptstyle\pm5.9}$ & $\mathbf{139.7}{\scriptstyle\pm9.3}$ \\
    \method{} + MPPI & $166.1{\scriptstyle\pm11.4}$ & $231.7{\scriptstyle\pm16.2}$ & $0.019{\scriptstyle\pm0.005}$ & $0.104{\scriptstyle\pm0.021}$ & $143.2{\scriptstyle\pm8.9}$ & $124.6{\scriptstyle\pm9.2}$ \\
    \bottomrule
  \end{tabularx}
\end{table}

The CEM branch attains higher ID and OOD return and lower rollout MSE. Because MPPI also changes the data-collection and training distributions, this measures the end-to-end effect of the planner within \method{}.

\section{Additional Results and Supplementary Analysis}\label{app:additional}

\subsection{Hamiltonian Energy, Geometry, and Control Coupling}\label{app:diagnostics}

\begin{figure}[tb]
  \centering
  \begin{subfigure}[t]{0.245\linewidth}\centering
    \includegraphics[width=\linewidth]{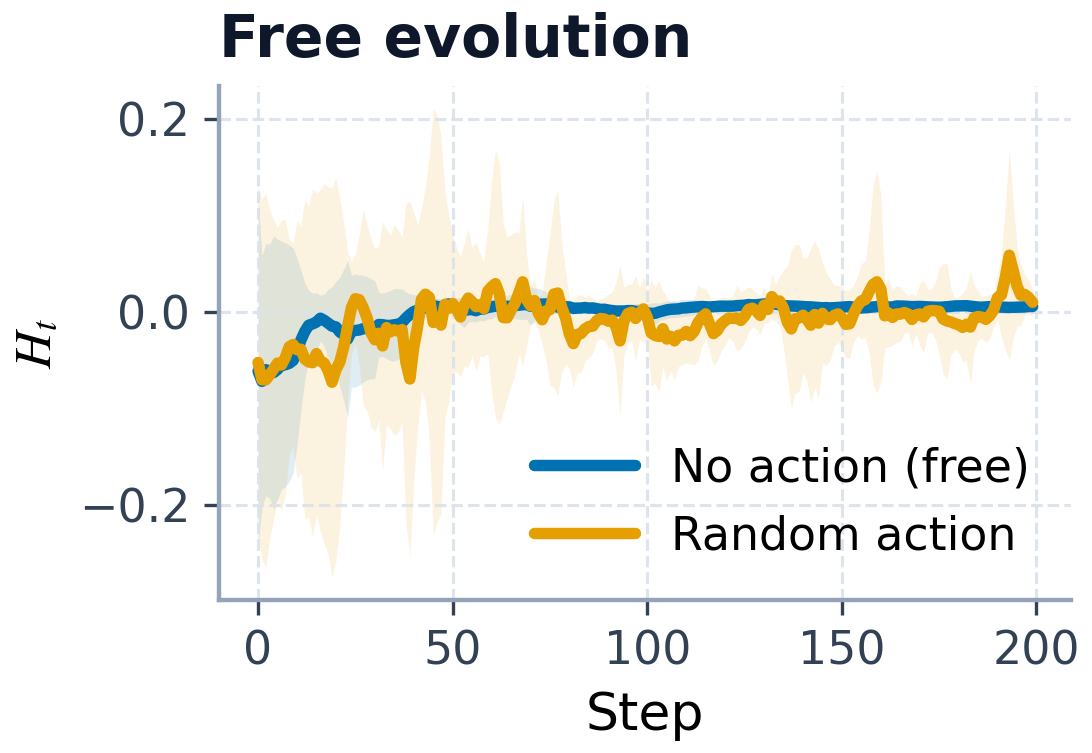}
    \caption{\footnotesize Free evolution}\label{fig:mech-h-freerun-cheetah}
  \end{subfigure}\hfill
  \begin{subfigure}[t]{0.245\linewidth}\centering
    \includegraphics[width=\linewidth]{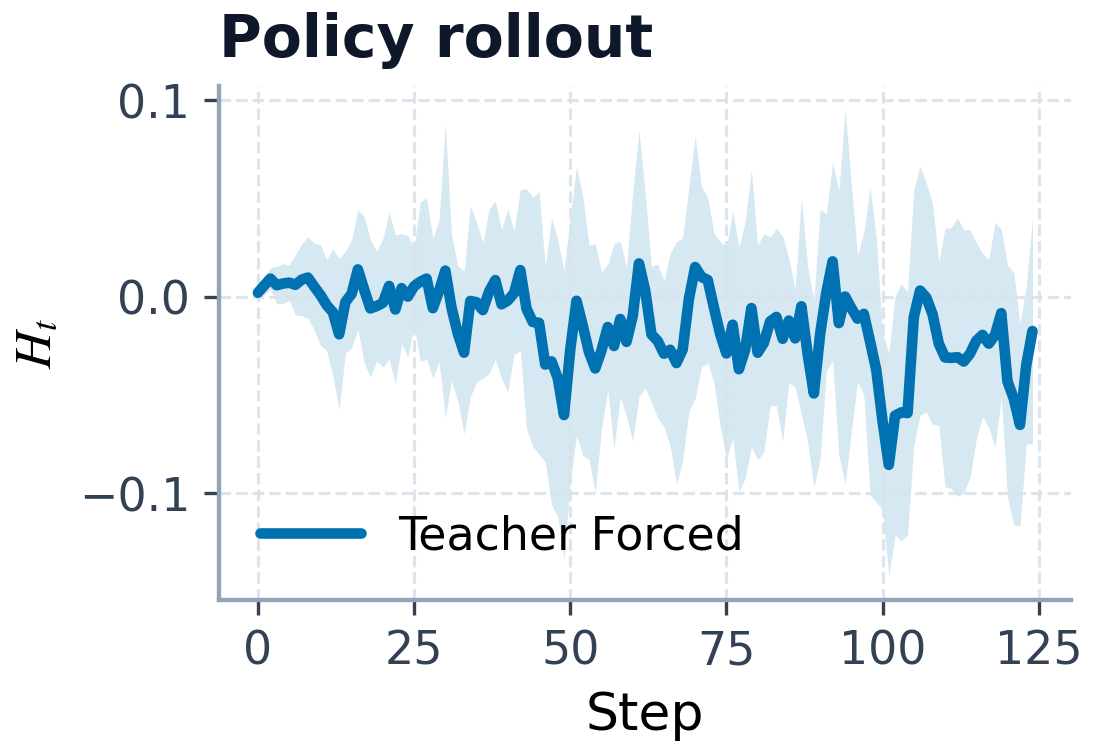}
    \caption{\footnotesize Policy rollout}\label{fig:mech-h-tf-cheetah}
  \end{subfigure}\hfill
  \begin{subfigure}[t]{0.245\linewidth}\centering
    \includegraphics[width=\linewidth]{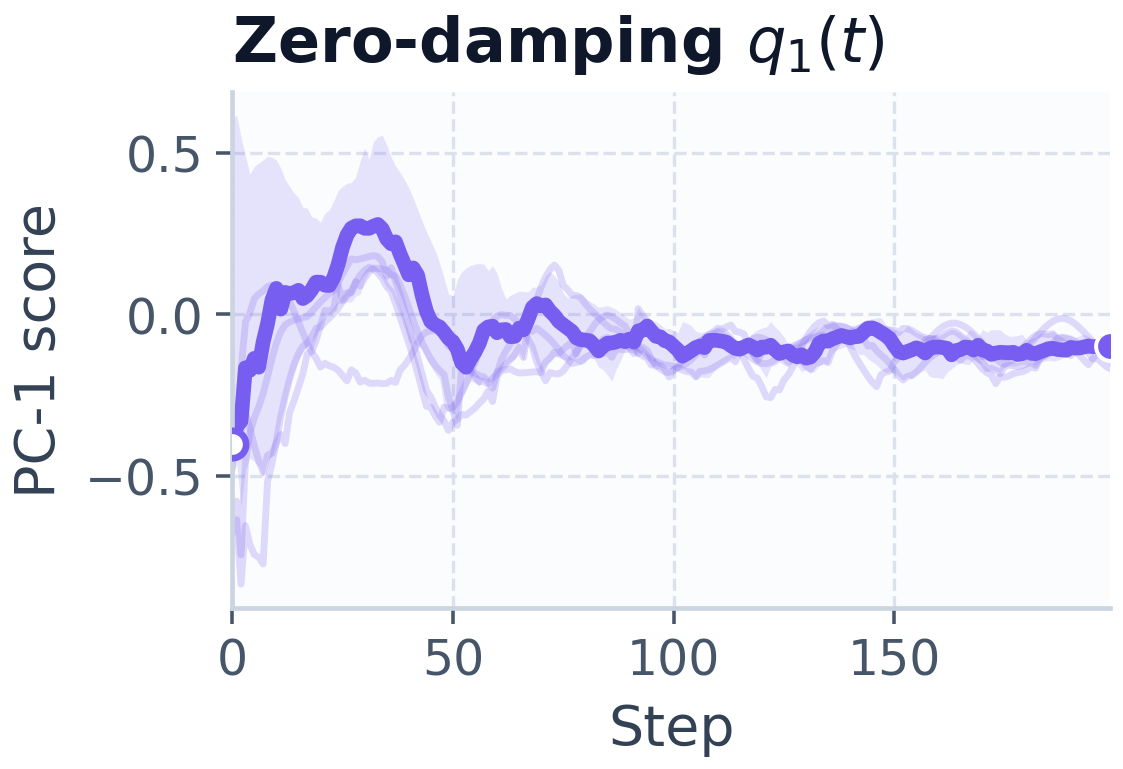}
    \caption{\footnotesize Zero-damping $\tilde q_1(t)$}\label{fig:phase-cheetah-q}
  \end{subfigure}\hfill
  \begin{subfigure}[t]{0.245\linewidth}\centering
    \includegraphics[width=\linewidth]{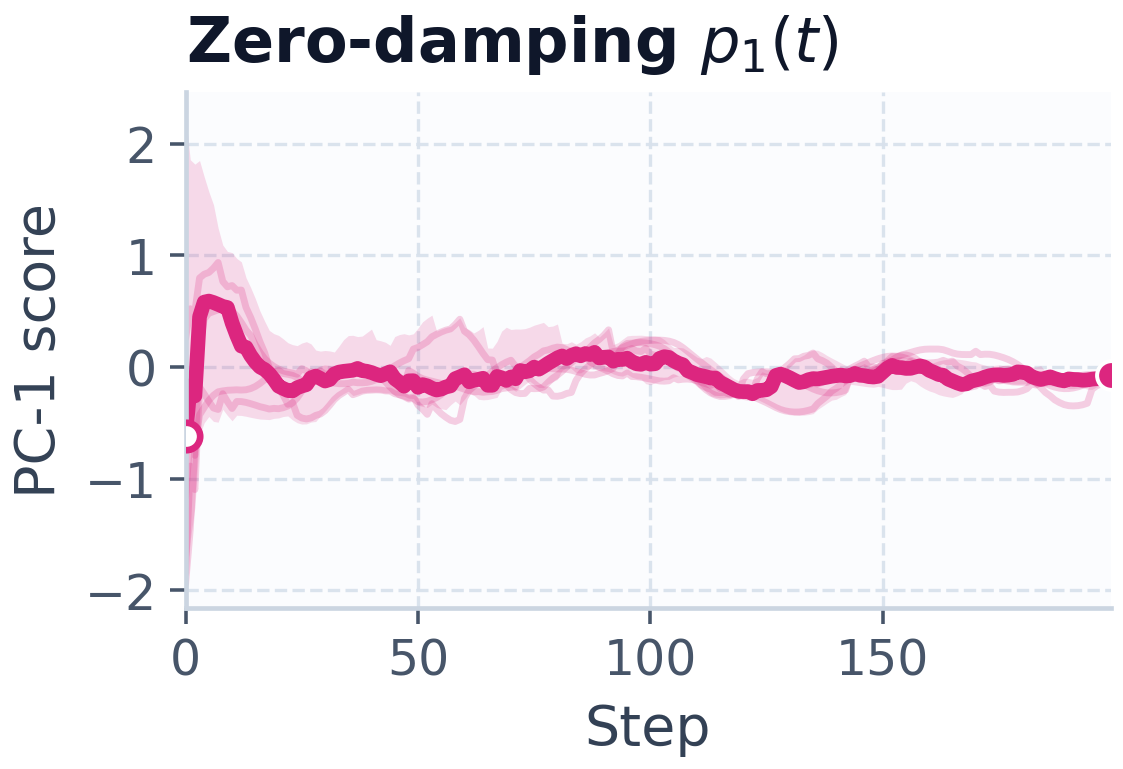}
    \caption{\footnotesize Zero-damping $\tilde p_1(t)$}\label{fig:phase-cheetah-p}
  \end{subfigure}
  \caption{Cheetah Run Hamiltonian diagnostics. \textbf{(a)} Hamiltonian energy under zero external force versus random action; \textbf{(b)} energy under policy rollout; \textbf{(c,d)} the two zero-damping Hamiltonian traces $\tilde q_1(t)$ and $\tilde p_1(t)$. Cheetah Run is used as the representative display.}
  \label{fig:h-panels}
  \label{fig:cheetah-geom}
\end{figure}

\begin{figure}[tb]
  \centering
  \includegraphics[width=0.36\linewidth]{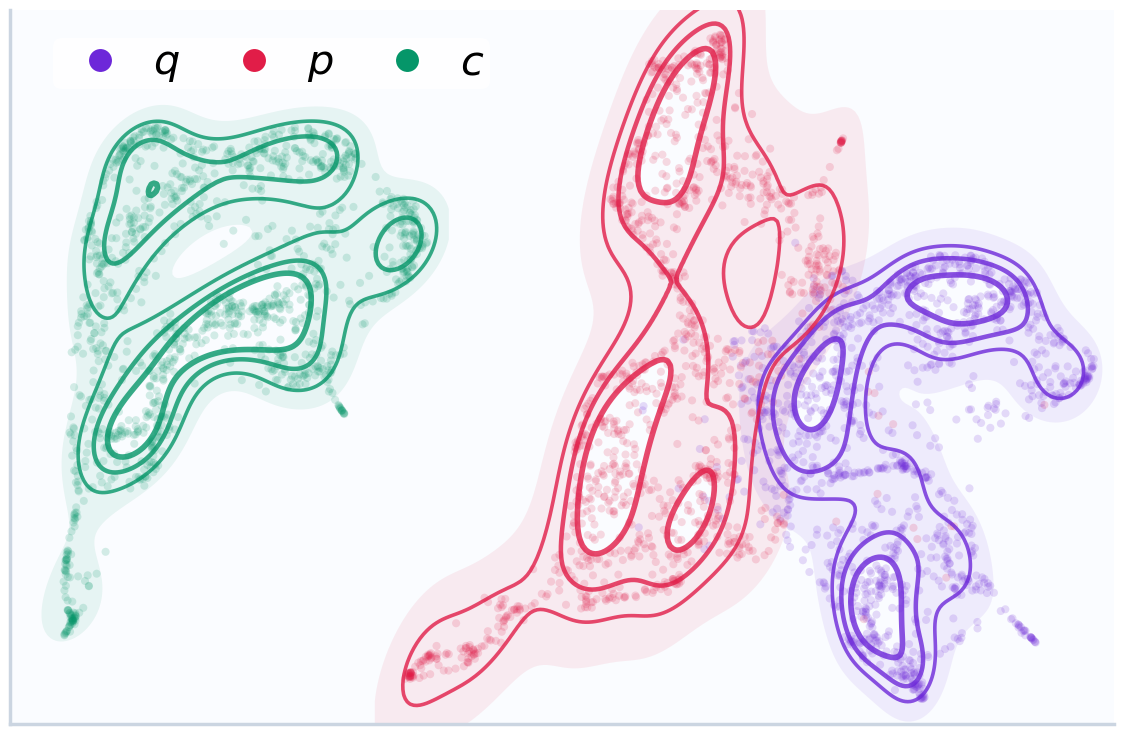}
  \caption{UMAP of the Cheetah Run latent coordinates: $\qv$ and $\pv$ form elongated manifolds with partial overlap, while $\cv$ stays a more compact context cloud.}
  \label{fig:cheetah-umap}
\end{figure}

\begin{figure}[tb]
  \centering
  \begin{subfigure}[t]{0.32\linewidth}\centering
    \includegraphics[width=\linewidth]{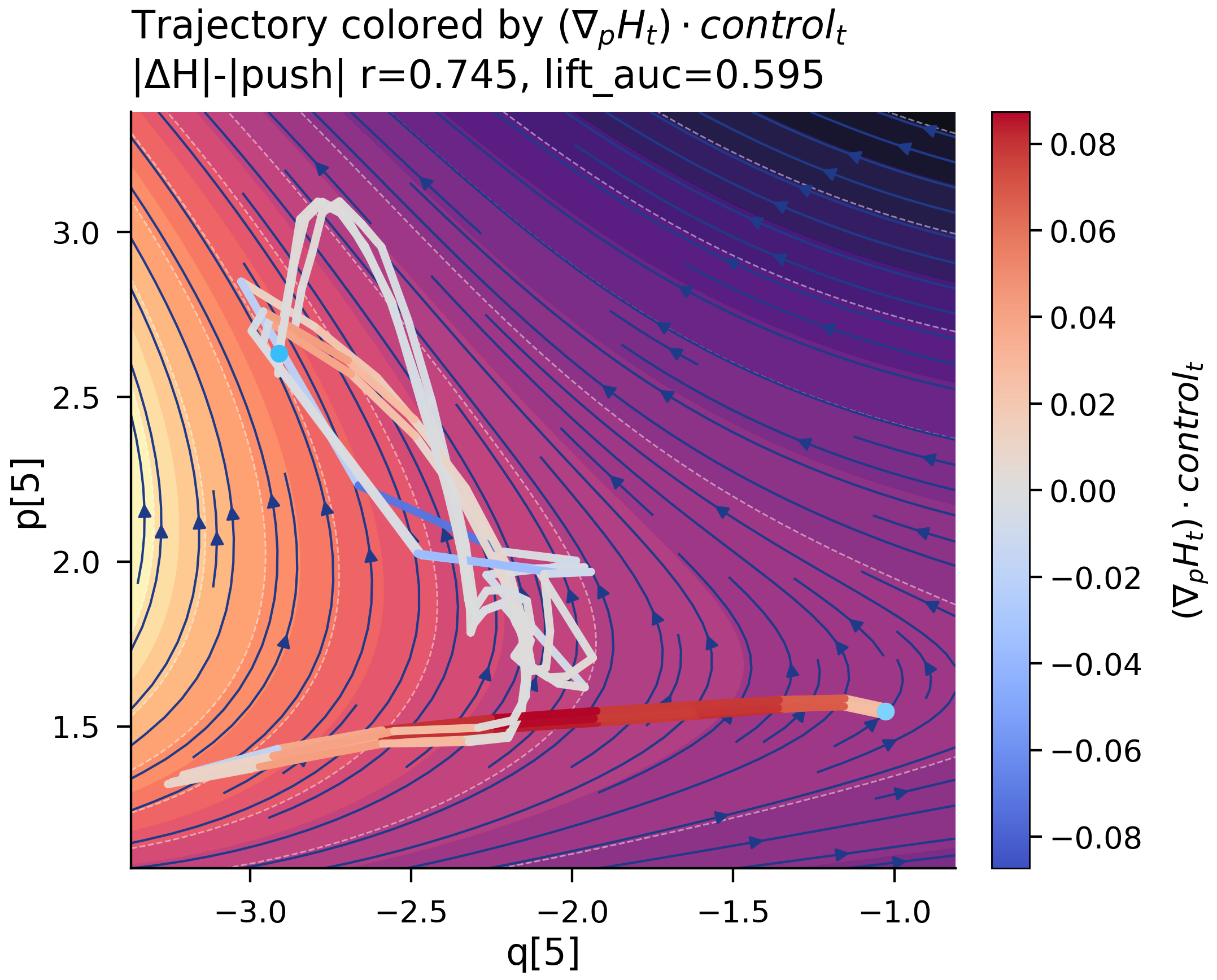}
    \caption{\scriptsize $i{=}j{=}5$: streamline alignment}\label{fig:hcp-flow-q5p5}
  \end{subfigure}\hfill
  \begin{subfigure}[t]{0.32\linewidth}\centering
    \includegraphics[width=\linewidth]{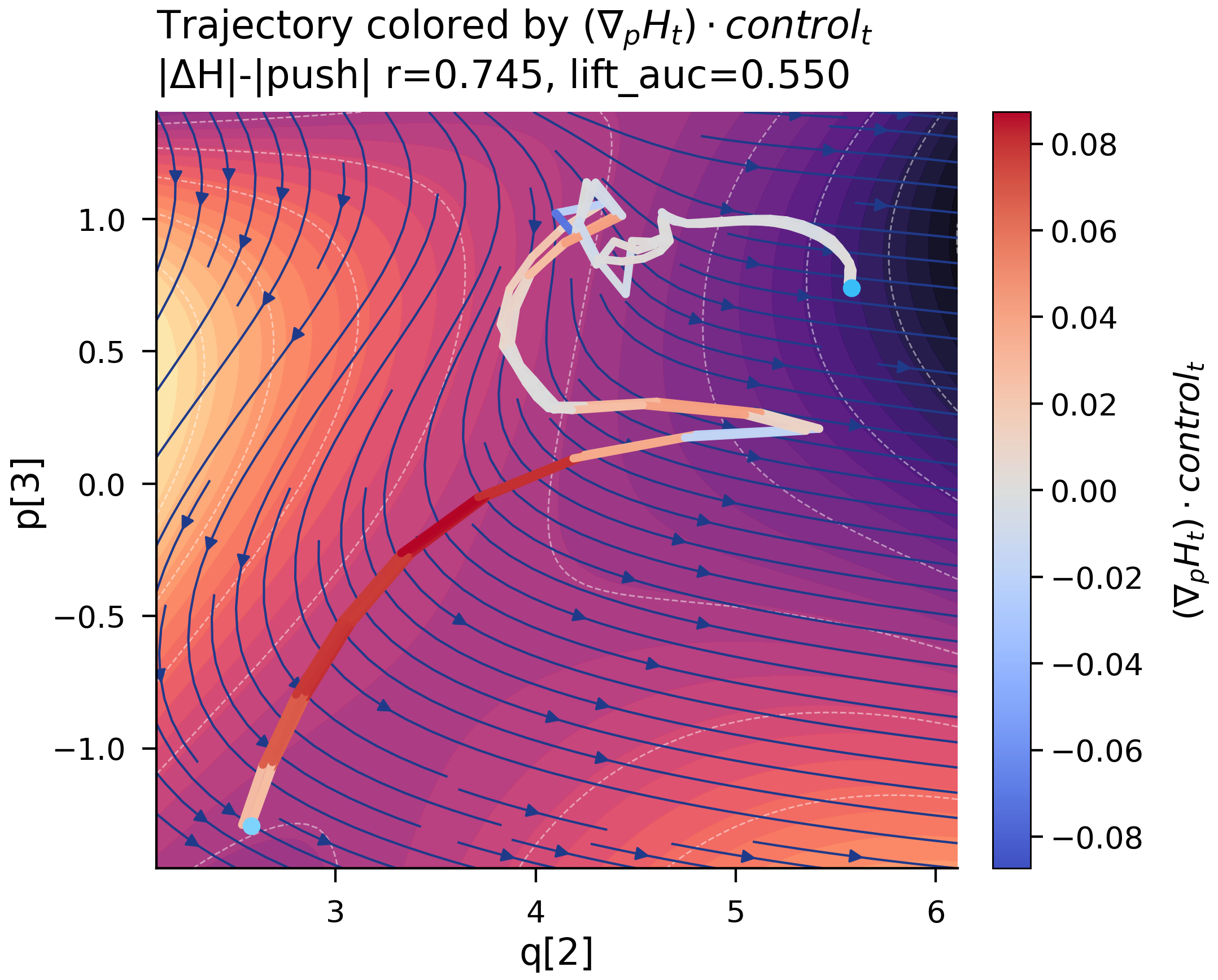}
    \caption{\scriptsize $(q_2,p_3)$, $i{\ne}j$: deviation}\label{fig:hcp-flow-q2p3}
  \end{subfigure}\hfill
  \begin{subfigure}[t]{0.32\linewidth}\centering
    \includegraphics[width=\linewidth]{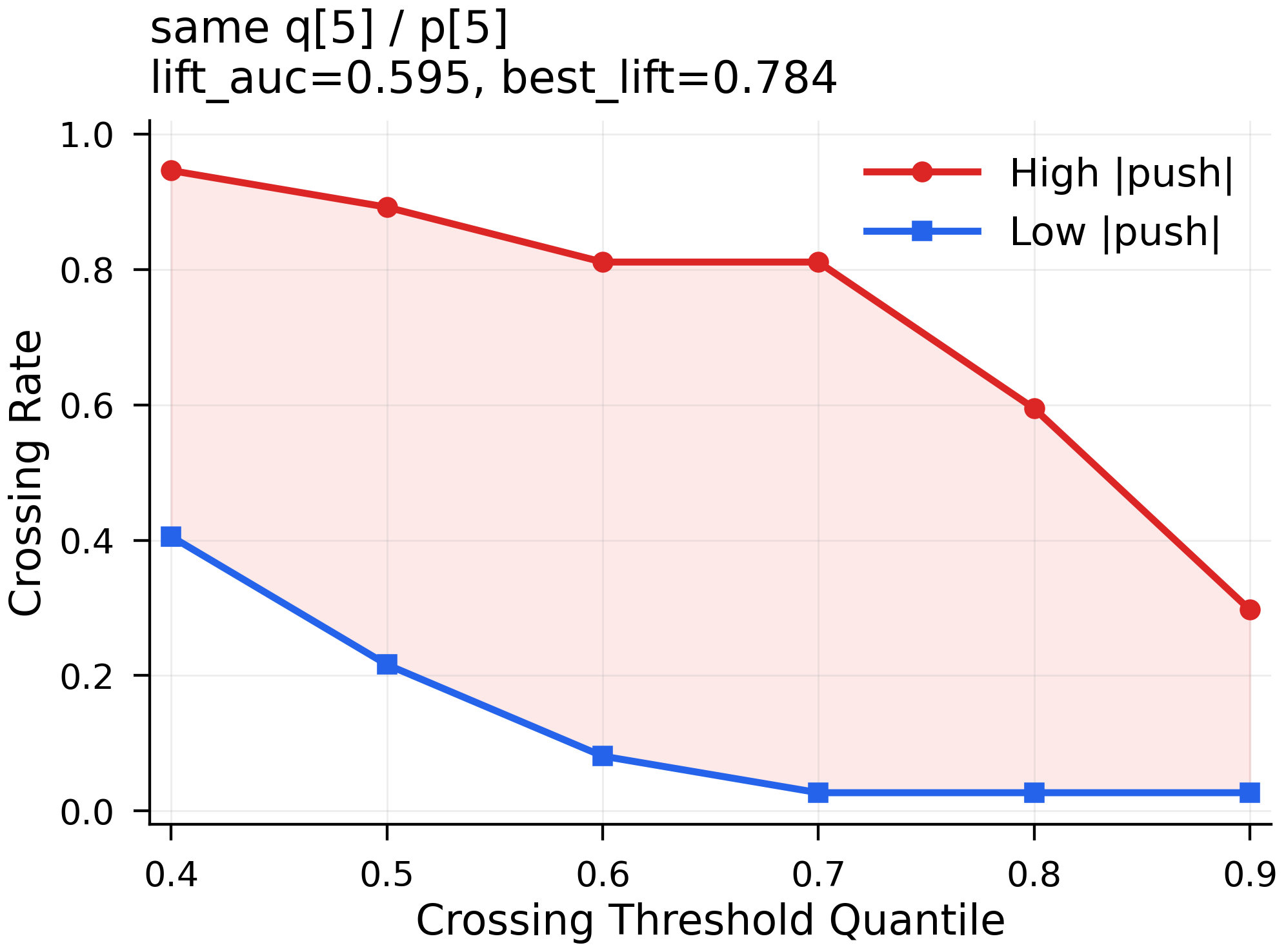}
    \caption{\scriptsize High vs.\ Low $|\mathrm{push}|$ crossing rate}\label{fig:hcp-crossing-q5p5}
  \end{subfigure}
  \caption{$\Ham$ contours and control coupling. The same diagnostic explains how external control moves rollouts across energy layers: high $|\mathrm{push}|=|\nabla_p\Ham\cdot\mathrm{control}|$ consistently yields larger equipotential crossing rates across thresholds.}
  \label{fig:hcp-control}
\end{figure}

This appendix gives the full mechanism diagnostics summarized in Section~\ref{ssec:mech}. Figures~\ref{fig:h-panels}--\ref{fig:hcp-control} collect all panels; the discussion in paragraphs (A)--(F) then refers back to them.

\paragraph{(A) The energy network is approximately stable and $\Ham$ matches the design expectation.}
The displayed mechanism analysis uses Cheetah Run as the running example. After reset, we inject uniformly random joint angular velocities in the range $\pm5$\,rad/s, set joint damping to $0$, and let the system freely evolve for $200$ steps; each regime contains $10$ episodes, with solid curves denoting the mean and shaded bands denoting $\pm$std across episodes. Figure~\ref{fig:mech-h-freerun-cheetah} compares the same $\Ham$ head under no-action and random-action conditions. The no-action curve remains nearly flat within episodes (about $1\%$ drift on Cheetah Run), with a narrow band across initial states. After random actions are introduced, the curve rises and widens, matching the physical expectation that external action continuously injects work into the system. This contrast is the key check: the learned scalar behaves differently under passive evolution and action-driven work, even though training imposes no hard energy conservation --- consistent with Section~\ref{sec:theory}, where action-free drift is expected at the percentage level.

\paragraph{(B) Policy rollouts show controlled energy variation.}
Figure~\ref{fig:mech-h-tf-cheetah} tracks $H_t$ along policy teacher-forced rollouts from $10$ different initial states. Policy execution continually applies control, so a flat conservation curve is not expected; instead $H_t$ tracks the ongoing behavior without abrupt jumps, with a bounded band across initial states. Together with the first panel this separates passive energy stability from policy-induced modulation, consistent with Table~\ref{tab:overview}: the q/p dynamics can vary under control while keeping finite-horizon error low.

\paragraph{(C) Geometry of the canonical and context subspaces.}

Figures~\ref{fig:phase-cheetah-q} and~\ref{fig:phase-cheetah-p} probe the Cheetah Run canonical traces behind the energy behavior. In the no-action, zero-damping setting, the randomized kick first produces a transient, but the uncontrolled Cheetah soon settles into a low-motion grounded posture; accordingly, $\tilde q_1(t)$ and $\tilde p_1(t)$ decay to small variations instead of drifting away, matching the passive rollout. The UMAP projection in Figure~\ref{fig:cheetah-umap} gives a complementary view of the same latent organization: $\qv$ and $\pv$ form elongated manifolds with partial overlap, while $\cv$ stays as a more compact context cloud. $\qv$ and $\pv$ overlap partially, as expected for coordinates sampled from one trajectory manifold, but remain distinct from each other and from $\cv$.

\paragraph{(D) Visualizing Hamiltonian geometry and control coupling.}
Figure~\ref{fig:hcp-flow-q5p5} visualizes the learned Hamiltonian on a canonical slice $(q_5,p_5)$: color denotes $\Ham(q,p)$, white curves are energy contours, and blue streamlines show $(\dot q,\dot p)=(\partial\Ham/\partial p,-\partial\Ham/\partial q)$. Teacher-forced rollout projections are colored by $\mathrm{push}_t=(\nabla_p\Ham_t)\cdot\mathrm{control}_t$. Local directions often follow the Hamiltonian flow; small $|\mathrm{push}|$ steps tend to move tangentially along contours, whereas large $|\mathrm{push}|$ steps more often cross energy layers. This is exactly the controlled-Hamiltonian story: tangential motion preserves the learned energy level locally, while control projected onto $\nabla_p\Ham$ performs work and changes energy. The quantitative correlations support the visual pattern, with $\mathrm{corr}(\mathrm{sign}\,\Delta\Ham,\mathrm{push})=0.757$ and $\mathrm{corr}(|\Delta\Ham|,|\mathrm{push}|)=0.745$.

\paragraph{(E) Non-conjugate slice.}
Figure~\ref{fig:hcp-flow-q2p3} projects the same rollout onto the non-conjugate slice $(q_2,p_3)$. Because this is not a matched canonical pair, the projected streamlines are not expected to align; the panel checks that the control--energy association is not specific to one slice. The correlations $\mathrm{corr}(\mathrm{sign}\,\Delta\Ham,\mathrm{push})=0.757$ and $\mathrm{corr}(|\Delta\Ham|,|\mathrm{push}|)=0.745$ are computed on the full latent state and are therefore slice-independent; the slice-specific contour-crossing lift remains above chance on $(q_2,p_3)$ (lift\_auc $=0.550$) and is higher on the conjugate pair ($0.595$), as the canonical structure predicts.

\paragraph{(F) High/low $|\mathrm{push}|$ contour-crossing rates.}
Figure~\ref{fig:hcp-crossing-q5p5} quantifies this effect by thresholding contour crossings and splitting timesteps by $|\mathrm{push}|$. The threshold sweep avoids relying on a single contour resolution. As the threshold increases, both crossing rates decrease, but the High-$|\mathrm{push}|$ curve stays above the Low-$|\mathrm{push}|$ curve (lift\_auc $=0.595$, best lift $=0.784$), confirming that stronger control projection along $\nabla_p\Ham$ makes contour crossings more likely.

\subsection{Additional Tasks: Walker Run and Hopper Hop}\label{app:extra-tasks}

The four core tasks underrepresent sustained coordination, momentum, and repeated contact, so we add two tasks under the identical protocol (100k env steps, seeds 7/8/9, same budget and evaluation pipeline). Learning curves are not reproduced for these two tasks; we report final return and the Avg.\ AUC pooled over all seed-runs of the core suite and of all six tasks (AUC computed as in Table~\ref{tab:overview}).

\begin{table}[H]
  \centering
  \caption{Additional tasks (mean$\pm$std, 3 seeds), same protocol and budget as Table~\ref{tab:overview}: final return at 100k env steps, and the Avg.\ AUC pooled over all seed-runs of the core suite (Table~\ref{tab:overview}) and of all six tasks.}
  \label{tab:extra-tasks}
  \footnotesize
  \setlength{\tabcolsep}{4pt}
  \begin{tabularx}{\linewidth}{@{}l*{5}{>{\centering\arraybackslash}X}@{}}
    \toprule
    \headrow Final return @ 100k $\uparrow$ & \method{} & \baselinetdmpc{} & \baselinedreamer{} & \baselinesac{} & \baselineppo{} \\
    \midrule
    Walker Run & $\mathbf{151.99}{\scriptstyle\pm33.27}$ & $101.85{\scriptstyle\pm29.72}$ & $\underline{104.57}{\scriptstyle\pm22.93}$ & $56.54{\scriptstyle\pm21.08}$ & $19.13{\scriptstyle\pm3.16}$ \\
    Hopper Hop & $\mathbf{34.46}{\scriptstyle\pm8.93}$ & $13.97{\scriptstyle\pm7.76}$ & $11.30{\scriptstyle\pm0.27}$ & $\underline{14.70}{\scriptstyle\pm1.22}$ & $4.70{\scriptstyle\pm0.21}$ \\
    \midrule
    Avg.\ AUC $\uparrow$, core suite (4 tasks) & $\mathbf{117.9}$ & $\underline{107.7}$ & $58.6$ & $70.9$ & $30.3$ \\
    Avg.\ AUC $\uparrow$, all six tasks & $\mathbf{91.9}$ & $\underline{83.3}$ & $51.7$ & $52.8$ & $22.7$ \\
    \bottomrule
  \end{tabularx}
\end{table}

\method{} is best on both, with gains over \baselinetdmpc{} of $+49\%$ on Walker Run and $+147\%$ on Hopper Hop. Combined with Table~\ref{tab:overview}, this gives first place on four of six tasks and second on the remaining two. Pooled over all 18 seed-runs per method, the six-task Avg.\ AUC is $91.9$ against $83.3$ for \baselinetdmpc{} ($+10.3\%$), so the core-suite sample-efficiency advantage carries over. These results extend the evaluation to control requiring sustained coordination and repeated contact, supporting the combined memory and geometric design.

\end{document}